\documentclass[preprint,review,3p,12pt,times]{elsarticle}

\usepackage{amssymb}

\usepackage{amsthm} 
\usepackage{lineno} 

\usepackage[utf8]{inputenc} 
\usepackage[T1]{fontenc}    
\usepackage{url}            
\usepackage{xspace}         
\usepackage{xcolor}         

\usepackage[unicode]{hyperref}
\usepackage{setspace}
\usepackage{caption}
\usepackage[most]{tcolorbox}

\usepackage{amsmath}
\usepackage{amssymb}
\usepackage{amsfonts}       

\usepackage{bbm}

\usepackage{graphicx}       
\graphicspath{{figures/}}   
\usepackage{subfigure}
\usepackage{booktabs}       
\usepackage{multirow}
\usepackage{colortbl} 
\usepackage{wrapfig}
\usepackage{adjustbox}
\usepackage{nicefrac}       
\usepackage{enumitem}
\usepackage{bbm}
\usepackage{listings}
\usepackage{courier}
\usepackage{tcolorbox}
\usepackage{capt-of}
\usepackage[normalem]{ulem}

\newcommand{\newtexttt}[1]{\lstinline+#1+}
\usepackage[dvipsnames,svgnames,x11names]{xcolor} 
\usepackage{hyperref}
\usepackage[capitalize,nameinlink]{cleveref}
\crefname{figure}{Fig.}{Figs.}
\crefname{section}{Sec.}{Secs.}
\Crefname{section}{Section}{Sections}
\crefname{table}{Tab.}{Tabs.}
\Crefname{table}{Table}{Tables}

\usepackage{acronym}
\usepackage[misc]{ifsym}

\makeatother

\usepackage{xcolor}
\newcommand{\pos}[1]{\textcolor{green}{\ensuremath{+\,#1}}}
\newcommand{\negnum}[1]{\textcolor{black!30}{\ensuremath{-\,#1}}}

\usepackage[table]{xcolor}
\usepackage{nicematrix}
\definecolor{rowgray}{RGB}{235,235,235} 

\journal{Pattern Recognition}

\begin{document}

\begin{frontmatter}



\title{Adaptive Chain-of-Focus Reasoning via Dynamic Visual Search and Zooming for Efficient VLMs}
\author[instA,instB]{Xintong Zhang\textsuperscript{*}} 
\author[instA,instB,instC,instD]{Zhi Gao\textsuperscript{*}} 
\author[instB]{Bofei Zhang}
\author[instA,instB]{Pengxiang Li}
\author[instB]{Xiaowen Zhang}
\author[instB]{Yang Liu}
\author[instB]{Tao Yuan}
\author[instA,instD]{Yuwei Wu\textsuperscript{\dag}} 
\author[instD]{Yunde Jia}
\author[instB,instC,instE]{Song-Chun Zhu}
\author[instB]{Qing Li\textsuperscript{\dag}} 

\affiliation[instA]{
  organization={School of Computer Science \& Technology, Beijing Institute of Technology},
  city={Beijing},
  country={China}
}
\affiliation[instB]{
  organization={State Key Laboratory of General Artificial Intelligence, BIGAI},
  city={Beijing},
  country={China}
}
\affiliation[instC]{
  organization={School of Intelligence Science and Technology, Peking University},
  city={Beijing},
  country={China}
}
\affiliation[instD]{
  organization={Guangdong Laboratory of Machine Perception and Intelligent Computing, Shenzhen MSU--BIT University},
  city={Shenzhen},
  country={China}
}
\affiliation[instE]{
  organization={Department of Automation, Tsinghua University},
  city={Beijing},
  country={China}
}




\tnotetext[projectpage]{Project page: \href{https://cof-reasoning.github.io/}{\texttt{cof-reasoning.github.io}}}
\nonumnote{\textsuperscript{*}\;These authors contributed equally to this work.}
\nonumnote{\textsuperscript{\dag}\;Corresponding authors:
\href{mailto:yuwei.wu@bit.edu.cn}{yuwei.wu@bit.edu.cn};
\href{mailto:liqing@bigai.ai}{liqing@bigai.ai}.}



\begin{abstract}

Vision Language Models (VLMs) have achieved remarkable success, particularly with "think-with-image" paradigms that enhance reasoning by actively image zooming to explore visual details, moving beyond reliance on purely textual thought processes. However, this approach presents a challenge in balancing performance with efficiency, as proactive zooming incurs massive computational costs and may impair global understanding. To address this problem, we introduce adaptive chain-of-focus (Adaptive-CoF), a framework that teaches VLMs to perform visual search and zooming only when necessary, based on obtained visual cues and the given questions, achieving efficient multimodal reasoning. We enable this capability through a two-stage pipeline: (1) supervised fine-tuning on an introduced MM-Adaptive-CoF SFT dataset that is constructed by a visual search agent with multi-step reasoning trajectories under diverse resolutions and question complexities, and (2) reinforcement learning with an adaptive group-aware reward (AGAR) on MM-Adaptive-CoF RL dataset, allowing the model to master an adaptive strategy. Our experiments show Adaptive-CoF achieves superior performance with exceptional efficiency. On the $V^\star$ benchmark, it reduces zoom-in operations by 75\% compared to proactive models and achieves comparable even better accuracy with nearly 50\% fewer tokens, establishing a new paradigm for efficient and accurate VLMs.
The code is available at \url{https://github.com/xtong-zhang/Chain-of-Focus}.

\end{abstract}




\begin{keyword}


Vision-Language Model \sep Multimodal Reasoning  \sep   Supervised Fine-tuning \sep Reinforcement Learning \sep Think-with-Image \sep Adaptive Reasoning
\end{keyword}

\end{frontmatter}



\section{Introduction}
Recent Vision-Language Models (VLMs), which couple visual encoders with large language models (LLMs)~\cite{openai2023gpt4v,zhu2023minigpt,yin2023survey}, have shown substantial progress. A key challenge toward more powerful and general-purpose VLMs lies in developing robust multimodal reasoning capabilities.
Early multimodal reasoning efforts primarily focused on the textual space, where a model generates a chain of thought using only language after an initial and holistic perception of the image~\cite{zhang2025r1,yang2025r1,wei2022chain,zhang2022automatic,gaomulti}. While effective for general understanding, this approach struggles with tasks requiring fine-grained information, particularly when dealing with small objects in high-resolution images, as crucial visual details may be lost or compressed in the one-time visual encoding~\cite{ke2025language}.
A key milestone in overcoming this limitation is the "think with image" interactive paradigm, which is inspired by the long reasoning breakthroughs like OpenAI-o1~\cite{jaech2024openai} and DeepSeek-R1~\cite{guo2025deepseek}. The "think with image" paradigm, notably demonstrated by visual grounded reasoning models like OpenAI's o3~\cite{openai2025o3}, which can actively explore visual information by interleaving textual thoughts with new visual evidence from dynamically manipulated image regions. 
Specifically, the model can ground and zoom in on relevant image regions during reasoning, allowing it to acquire fine-grained visual cues.
Building on this interactive paradigm, subsequent visual grounded reasoning models like DeepEyes~\cite{zheng2025deepeyes} and Pixel Reasoner~\cite{su2025pixel}, have further leveraged Reinforcement Learning (RL) to guide this process~\cite{Sarch2025GroundedRL,Wang2025VGR,Liu2025UniVGR1,Qiu2025GRIT,Liu2025VisualToolAgent}. These works yield significant benefits, enabling models to perceive details otherwise invisible and to ground their reasoning in concrete visual evidence.

Despite the advantages, these methods reveal limitations. Constant zooming poses two critical issues: it is highly inefficient due to the generation of excessive visual tokens from unnecessary zooming operations, and it can be detrimental for the model by causing the model to lose in the long context. Therefore, an intelligent VLM should learn to adaptively and dynamically decide whether the current visual evidence is sufficient and when to zoom in for essential and fine-grained details, effectively balancing accuracy with efficiency.

\begin{figure}[htbp!]
\centering
\includegraphics[width=0.9\linewidth]{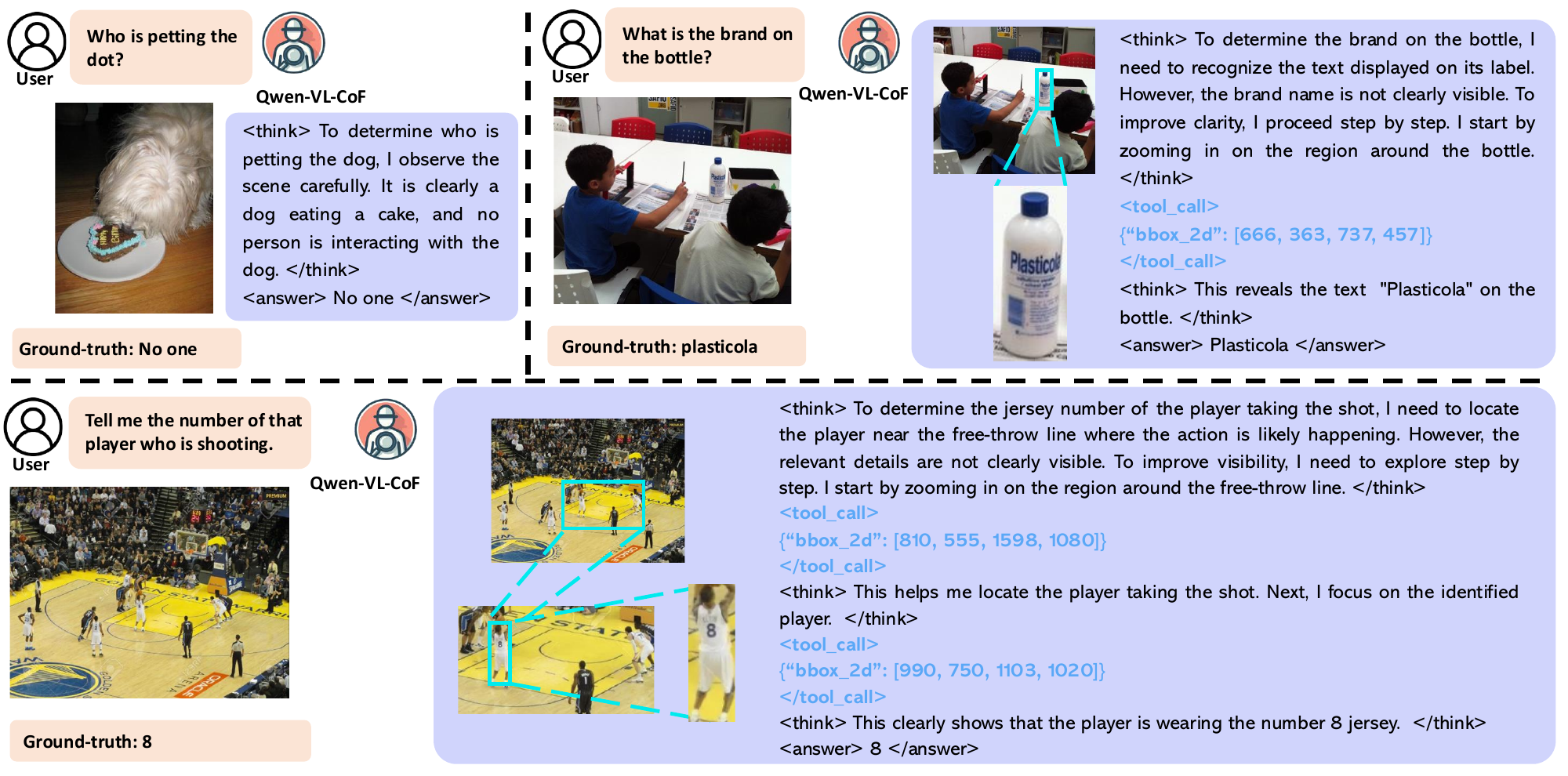}
\vskip -0.1in
    \caption{The proposed Adaptive-CoF method enables VLMs to perform adaptive search and zooming to obtain a necessary chain of visual information for answering. The VLMs answer the query if the visual information is sufficient; otherwise, it zooms in on key regions for more visual details.
}
\label{fig:illustration_examples}
\end{figure}

In this paper, we propose the Adaptive Chain-of-Focus (Adaptive-CoF) method that allows VLMs to perform adaptive search and zooming on key image regions (see~\cref{fig:illustration_examples} for details), thus creating a chain of focus steps for multimodal reasoning. For example, when the query can be answered from a global view, the model answers directly. When the query demands details from small regions, the model searches for and zooms in on key image regions to extract more visual cues. In implementation, the visual tokens corresponding to these key regions are appended to the context, allowing the VLM to gather new visual evidence and analyze the image more thoroughly, accurately, and reliably. Crucially, our method is adaptive: it only performs visual search and zooming when necessary, reducing computational costs while maintaining high performance.

To equip VLMs with this capability, we present a two-stage training pipeline, including supervised fine-tuning (SFT) and reinforcement learning (RL). In the SFT stage, we introduce the MM-Adaptive-CoF SFT dataset, a collection of 5K data with reasoning trajectories constructed from the SAM~\cite{kirillov2023segment} dataset across various image resolutions. We first synthesize tasks for these images, then deploy a visual agent with multiple tools to generate a reasoning trajectory to solve the task. After that, the trajectory is summarized into a coherent Adaptive-CoF reasoning process. We use this dataset to fine-tune a Qwen2.5-VL-7B model for a cold start. In the RL stage, we propose an adaptive group-aware reward (AGAR), leveraging outcome accuracies and formats to further refine the model’s strategy without human priors. This allows the model to learn adaptive reasoning when a direct answer is sufficient or when a zoom-in operation is necessary. The model is trained with 10,000 samples from the MM-Adaptive-CoF RL dataset, enabling effective reinforcement of both decision-making and visual grounding behaviors.

Experiments on multiple and challenging benchmarks show the effectiveness of Adaptive-CoF, showing significant improvements over its base model, Qwen2.5-VL-7B, with an 18.9\% gain on the V$^{\star}$ Bench and a notable 8.6\% boost on the MME-RealWorld-Lite benchmark. This demonstrates its superior capability in high-resolution perception, comprehensive reasoning, and hallucination reduction. Compared with visual grounded reasoning models such as DeepEyes~\cite{zheng2025deepeyes} and Pixel Reasoner~\cite{su2025pixel}, which perform proactive zooming, Adaptive-CoF achieves a superior balance between performance and efficiency. This is best illustrated on the demanding HR-Bench 4K benchmark, where Adaptive-CoF reduces zoom-in operations adaptively, cutting computational costs and using only about 5.4\% of the visual tokens compared to DeepEyes. This strategic reduction in zooming is consistent across benchmarks, where the number of zoom-in operations decreases by up to 77.8\% (on HR-Bench 4K vs. DeepEyes). Together, these results highlight Adaptive-CoF’s powerful combination of leading accuracy and exceptional efficiency.


Our main contributions are summarized as follows.
(1) We propose the Adaptive Chain-of-Focus (Adaptive-CoF) method, which substantially enhances multimodal reasoning in vision-language models by dynamically balancing reasoning accuracy and computational efficiency. 
(2) We introduce a data collection pipeline and produce Adaptive-CoF data via a visual search agent, leading to MM-Adaptive-CoF, a dataset containing 5K Adaptive-CoF samples across multiple domains, different image resolutions, and diverse queries.
(3) We develop an Adaptive-CoF model, an advanced VLM that could perform adaptive visual search and reasoning on images, leading to thorough, accurate, and reliable visual understanding.

\section{Related Work}

\subsection{Vision Language Models}
Developing powerful VLMs is a hot research topic in the multimodal learning community~\cite{liu2024visual}. 
Existing VLMs combine a visual encoder (\emph{e.g.}, ViT~\cite{dosovitskiy2020image}), an LLM (\emph{e.g.}, Qwen-2.5~\cite{yang2024qwen2}), and a projector (\emph{e.g.}, MLP or Q-Former~\cite{li2023blip}) that connects the visual encoder and LLM for multimodal understanding. 
The visual encoder encodes images into visual tokens, and the projector converts the visual tokens into the language space.
Finally, the visual tokens and textual tokens are
combined and fed into the LLM for autoregressive prediction.
Recent models, such as LLaVA-OneVision~\cite{li2024llava}, InternVL~\cite{chen2024internvl}, Qwen-VL~\cite{bai2023qwen}, and LLaVA-UHD~\cite{xu2024llava} have shown that using high-resolution images significantly improves the performance of visual perception and reasoning.
Compared to low-resolution images, high-resolution images contain more details, and processing high-resolution images usually requires more visual tokens, delivering more visual cues\cite{liu2024llavanext,hudson2019gqa}.
Different from existing methods that feed high-resolution images at first, our method performs adaptive search to identify and zoom in on key regions in a chain of focus, avoiding processing irrelevant regions for cost reduction and improving the reasoning capability of VLMs.

\subsection{Multimodal Reasoning}

Reasoning, a key mechanism in LLMs~\cite{wei2022chain,zhang2022automatic,gaomulti}, has increasingly extended to multimodal settings, driven by breakthroughs like OpenAI-o1~\cite{jaech2024openai} and DeepSeek-R1~\cite{guo2025deepseek}. 
A key milestone is OpenAI-o3~\cite{openai2025o3}, which proposes a think-with-image paradigm and integrates visual evidence via dynamic image manipulation (cropping and zooming) to enhance reasoning. Other research focuses on knowledge coordination~\cite{wang2024coordinating}, prompt engineering~\cite{hu2024prompting}, and improving foundational visual grounding~\cite{ke2025language}.
Building on this interactive paradigm, influential works have leveraged Reinforcement Learning (RL) to guide the visual reasoning process~\cite{Sarch2025GroundedRL,Wang2025VGR,Liu2025UniVGR1,Qiu2025GRIT,Liu2025VisualToolAgent}. Key approaches include DeepEyes~\cite{pezzotti2017deepeyes}, which encourages proactive visual exploration, and Pixel Reasoner~\cite{su2025pixel}, which employs a Curiosity-Driven Reasoning strategy. These methods use RL-driven zooming to interleave textual thoughts with new visual evidence, learning policies that decide where and when to zoom-in for fine-grained information. However, their explicit encouragement of proactive exploration results in frequent and redundant zoom-in operations, significantly increasing computational and token costs despite the gains in localized visual reasoning. To address these limitations, our proposed Adaptive-CoF enables adaptive reasoning, dynamically deciding whether to rely on current input or invoke additional visual operations only when necessary.

\subsection{LLM Adaptive Reasoning}
LLM adaptive reasoning aims to address the trade-off between efficiency and reasoning depth. Its core objective is to give models a dynamic capability to allocate computational resources based on task complexity. For simple queries, a model should respond concisely to save costs, while for complex problems, it should switch to a deeper, more computationally intensive reasoning mode.
Currently, two primary technical pathways are pursued to achieve this goal. The first involves constructing collaborative multi-model systems, where a lightweight "router" model pre-assesses and dispatches tasks to the most suitable expert model~\cite{ong2024routellm}. Alternatively, this approach can employ "speculative decoding," where a smaller model rapidly generates a draft response that is subsequently verified and refined by a more powerful one~\cite{liao2025reward}. 
The second pathway is to deploy multiple reasoning modes within a single, unified model, activating different operational states through specific prompt-based instructions~\cite{bercovich2025llama}~\cite{yang2025qwen3}~\cite{anthropic2025sonnet}.
However, existing methods often rely heavily on hand-crafted heuristic rules. A more promising direction is to develop a learning-based approach that enables the model to autonomously learn such judgments from data, automatically selecting the optimal reasoning path without manual directives.
In this work, we present a learning-based alternative for the multimodal domain. Rather than depending on external routers or pre-defined prompts, our framework leverages reinforcement learning to train the model to learn an internal, context-aware policy for deciding when deeper visual exploration is necessary.

\section{Method}

\subsection{Formulation}
Adaptive-CoF endows VLMs with adaptive multimodal reasoning in multiple steps,
by selectively invoking visual evidence via key region localization when necessary, or directly relying on textual reasoning when sufficient.
The $i$-th step is formulated as:
\begin{equation}
 \max \pi_{\theta} (r_i, o_i| {I}, q, h_i), 
  \label{eq:important}
\end{equation}
where $\pi_{\theta}$ denotes a VLM, ${I}$ is the input image, $q$ is the input query, and $h_i$ is the history. At the $i$-th step, $a_i$ and $o_i$ are the generated key regions and textual response, respectively. The key regions $r_i = \big[[x_1,y_1, x_2, y_2],\dots\big]$ contain the coordinates of bounding boxes. The textual response $o_i$ is the predicted answer if the regions are sufficient to answer the query $q$; otherwise, it is the intermediate reasoning content. The history $h_i$ includes the key regions ($r_j$), responses ($o_j$), and visual tokens ($t_j$) from all previous steps ($j < i$), and is formulated as $h_i=[r_1, o_1, t_1, \dots, r_{i-1}, o_{i-1}, t_{i-1}]$.

\subsection{Architecture}

\begin{figure}
    \centering
    \includegraphics[width=0.9\textwidth]{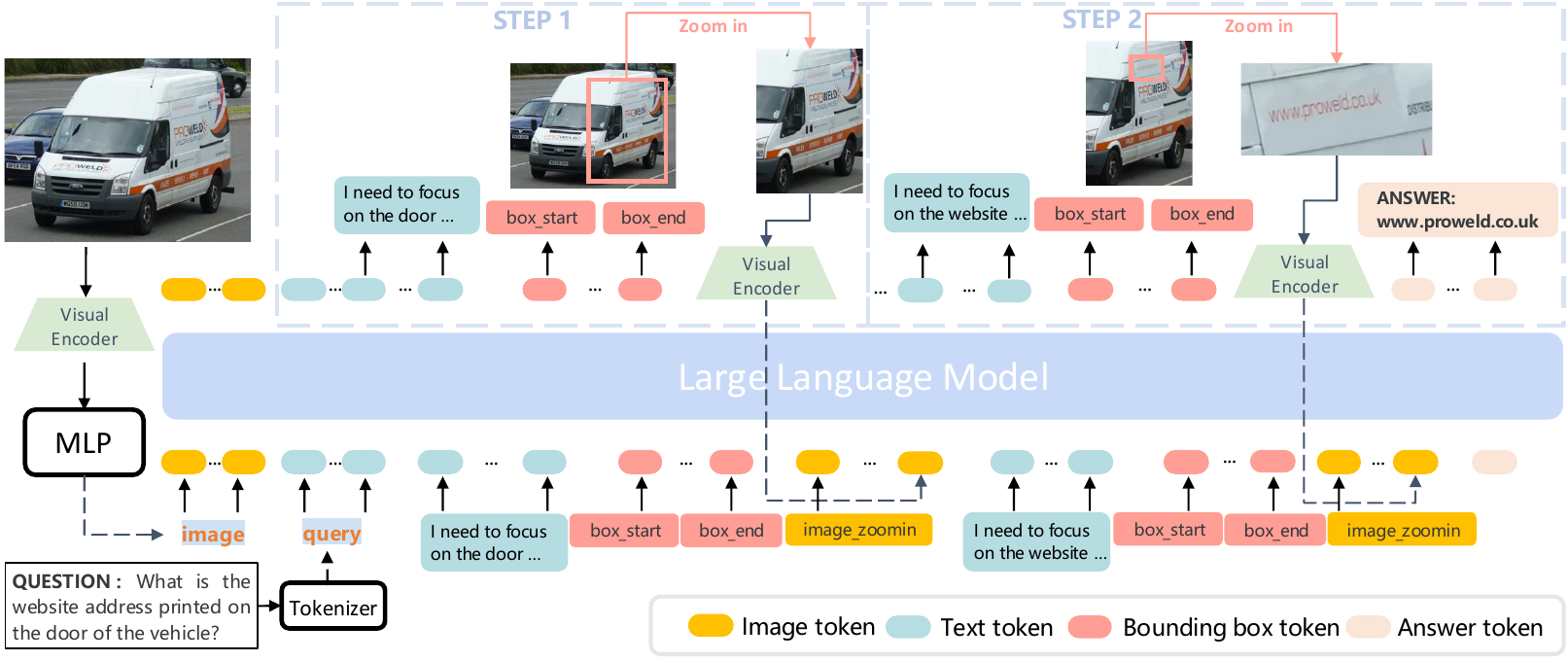}
  \vskip -0.1in
      \caption{Illustration of the Adaptive Chain-of-Focus framework. The textual tokens are in blue, visual tokens are in yellow, bounding box tokens are in red, and answer tokens are in orange. 
      }
\label{fig:chainoffocus}
\end{figure}

The overall process of Adaptive-CoF is shown in~\cref{fig:chainoffocus}, which is constructed on top of VLMs. 
The VLMs adopt the same architecture as commonly used VLMs (\emph{e.g.}, Qwen-VL), including a vision encoder, a projector, and an LLM. 
The visual encoder extracts visual tokens, the projector projects these visual tokens to the same space as textual tokens, and the LLM generates outputs based on visual tokens and textual tokens. 

At each reasoning step, the model may detect multiple key regions with bounding boxes, whose visual tokens are extracted and appended to the context. When the visual evidence is insufficient to directly answer the query, the model further identifies bounding boxes that provides closer inspection and applies zoom-in operations to enable fine-grained analysis. Each bounding box is denoted as $[x_1, y_1, x_2, y_2]$, where $(x_1, y_1)$ corresponds to the top-left corner of the region and $(x_2, y_2)$ corresponds to the bottom-right corner.

We utilize the tool calling capability of Qwen-2.5-VL for cropping key regions and zooming after the bounding boxes are generated, as shown in Fig.~\ref{fig:tool_reasoning_example}. This provides a standardized interface for issuing visual operations during reasoning. Once a tool call is generated, the system executes it in four steps: (i) cropping the indicated regions from the input image, (ii) enlarging them by a factor of two to expose finer-grained details, (iii) encoding the cropped regions using the visual encoder to obtain region-specific visual tokens, denoted as $o_j$ for the $j$-th step, and (iv) appending these visual tokens to the model context. This allows the model to jointly reason over both modalities. In this way, Adaptive-CoF not only detects candidate regions but also learns when and how to re-examine them through zoom-in operations.

\begin{figure}[t]
  \captionsetup{aboveskip=2pt,belowskip=0pt} 
  \centering
  \begin{tcolorbox}[
      width=.95\linewidth,      
      boxsep=2pt, left=4pt, right=4pt, top=3pt, bottom=3pt, 
      boxrule=0.3pt, arc=5pt,   
      before skip=0pt, after skip=0pt, enhanced
    ]
    \begingroup
    \setstretch{1}           
    \setlength{\parskip}{0pt}   
    \setlength{\parindent}{0pt} 
    \small

    \verb|<image>| Query: What is the website address printed on the door of the vehicle

    Think in the mind first, and then decide whether to call tools one or more times OR provide final answer. Format strictly as: \verb+<think>+...\verb+</think>+ \verb+<tool_call>+...\verb+</tool_call>+ (if any tools needed) OR \verb+<answer>+...\verb+</answer>+ (if no tools needed).

    \verb|<think>| To determine the website address printed on the door of the vehicle, I need to focus on the website text, but it is not clearly visible. To improve visibility, I proceed step by step. I first locate the door of the vehicle and obtain the bounding box $[x_1, y_1, x_2, y_2]$. \verb|</think>|

    \verb+<tool_call>+
\lstset{
  basicstyle=\ttfamily\footnotesize,
  aboveskip=2pt, belowskip=2pt,      
  columns=fullflexible, frame=none,
  showstringspaces=false, xleftmargin=0pt
}
\begin{lstlisting}
{ 
  "name": "image_zoom_in_tool", 
  "arguments": {"bbox_2d": [x1, y1, x2, y2]}
}
\end{lstlisting}
    \verb+</tool_call>+

    \verb|<image>|

    \verb|<think>|However, the website is still unclear. I further narrow the bounding box to focus specifically on the text area, yielding $[x_1, y_1, x_2, y_2]$. \verb|</think>|

    \verb+<tool_call>+
\begin{lstlisting}
{
  "name": "image_zoom_in_tool",
  "arguments": {"bbox_2d": [x1, y1, x2, y2]}
}
\end{lstlisting}
    \verb+</tool_call>+

    \verb|<image>|

    \verb|<think>| Now I can clearly read the text: www.proweld.co.uk. \verb|</think>|

    \verb|<answer>| www.proweld.co.uk \verb|</answer>|
    \endgroup
  \end{tcolorbox}
  \caption{An example of tool-based reasoning with bounding box generation and zoom-in operations.}
  \label{fig:tool_reasoning_example}
\end{figure}

\section{MM-Adaptive-CoF Dataset}

The MM-Adaptive-CoF dataset is meticulously constructed to facilitate the two-stage training of our Adaptive-CoF model, comprising distinct data splits tailored for supervised fine-tuning (SFT) and reinforcement learning (RL). It is built upon images sampled from the SA\_1B dataset~\cite{kirillov2023segment}, which predominantly features high-resolution images.

\subsection{Query-Response Generation}
\label{sub:qa_generation}

This initial stage focuses on generating high-quality query-response pairs, which form the foundational pool for constructing the MM-Adaptive-CoF dataset. We randomly select 360 images from the SA\_1B dataset and upscale them to a resolution of 4K to present sufficient visual details. For each image, we employ GPT-4.1 to generate four queries. To guarantee the reliability of the answers, we further require consistency between GPT-4.1 and Qwen-2.5-VL-72B: each query is answered by both models, and only those query–response pairs with the same answer are retained. This filtering process yields a set of high-quality QA pairs.

To explore different reasoning patterns, each image is resized to multiple resolutions (ranging from $\times224$ to $\times4\mathrm{K}$), and Qwen-2.5-VL-72B is prompted to assess whether the query can be directly answered at each resolution, along with providing an answer. This process is repeated five times per resolution, and only QA pairs with consistent judgments are retained. If the model consistently judges the query to be directly answerable and provides the correct answer, the pair was categorized as not requiring zoom-in; otherwise, it is considered to require zoom-in reasoning.
This meticulous process yields data that embodies two critical characteristics:
(1) Dynamic zoom-in requirements across resolutions. The same query can have dynamic zoom-in needs based on image resolution. For example, a query may require zooming in at low resolution but become directly answerable as visual detail increases. This pattern trains the model to adapt its reasoning to the available visual information.
(2) Diverse zoom-in needs per image. Within a single image, different queries have varying zoom requirements. Some can be answered from the overall context, while others demand intricate details accessible only through zooming. This trains the model to focus its attention based on the specific query.

\subsection{MM-Adaptive-CoF Supervised Fine-tuning (SFT) Dataset}

The data constructed during the query-response generation stage forms the foundation for the MM-Adaptive-CoF SFT dataset, specifically designed for Chain-of-Thought (CoT) training. From this initial pool, 2,201 QA pairs that necessitate zoom-in reasoning and 2,294 pairs that can be answered directly without zooming are selected.


To equip these samples with detailed Chain-of-Thought trajectories, we employ different generation strategies based on the reasoning path.
For QA pairs identified as requiring zoom-in reasoning, a visual search agent is further employed to generate corresponding step-by-step reasoning processes. For queries that can be answered directly without zooming in, GPT-4.1 is utilized to construct appropriate reasoning chains. This comprehensive approach ensures that the SFT dataset not only provides answers but also detailed explanatory reasoning, crucial for CoT training. We denote the resulting supervised fine-tuning dataset as $\mathbb{D}$, which serves as the training corpus for the SFT stage. The dataset's collection pipeline and its analytical breakdown provide a thorough understanding of its composition and utility.

\begin{figure}
    \centering
    \includegraphics[width=0.75\textwidth]{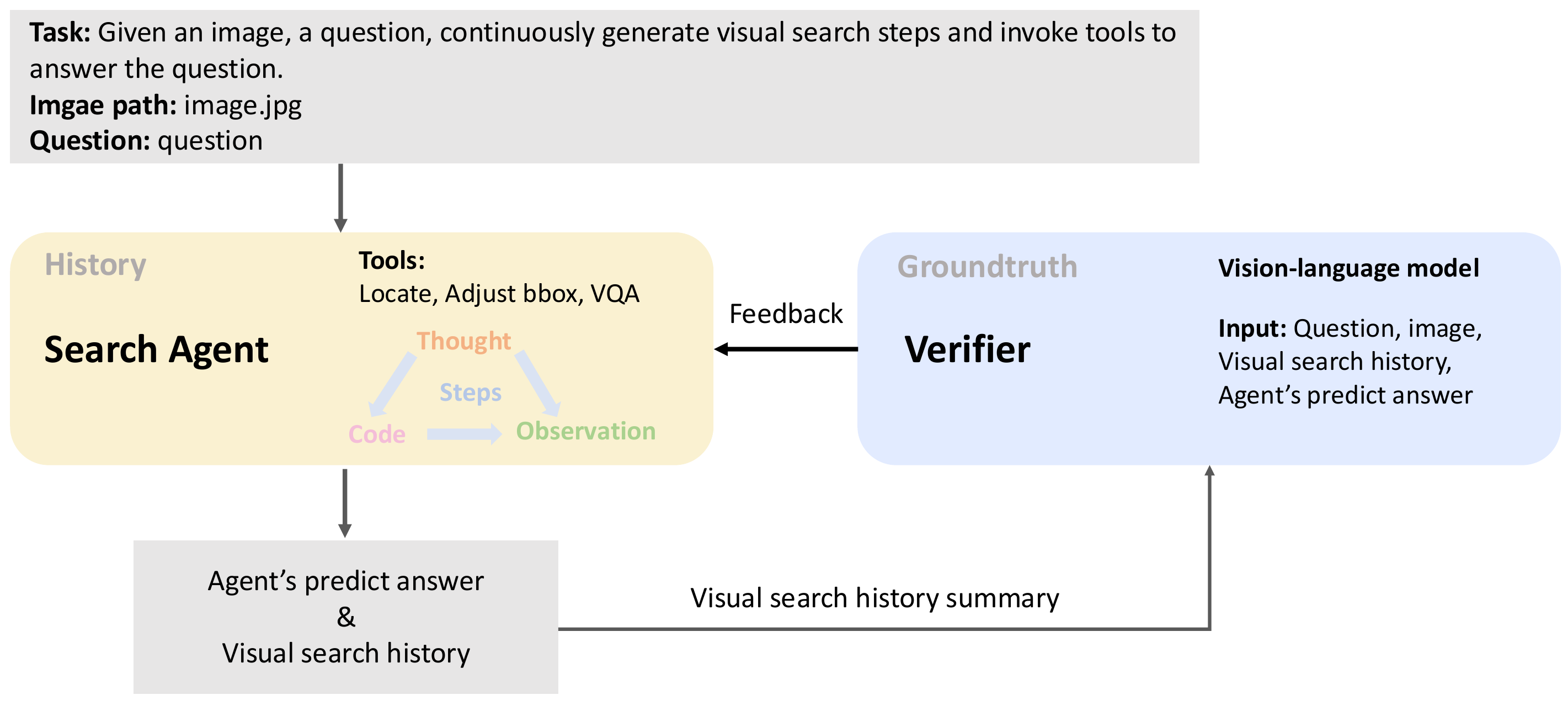}
      \caption{Architecture of the proposed visual search agent, consisting of the Search Agent and the Verifier.}
\label{fig:visualsearchagent}
\end{figure}

\subsubsection{Visual Search Agent}

We construct a visual search agent based on the ReAct framework~\cite{yao2023react}, with GPT-4.1-mini as its core controller. This architecture performs step-by-step reasoning by tightly coupling natural language thoughts with tool executions, as illustrated in ~\cref{fig:visualsearchagent}. The pivotal advantage of GPT-4.1-mini is its ability to directly perceive and analyze images, unlike models reliant solely on text-based descriptions. This allows it to generate more rational and efficient plans, making it exceptionally effective at coordinating the various visual modules and driving the agent's workflow.

To enable effective multimodal reasoning, the agent is equipped with four complementary components, all instantiated by Qwen2.5-VL-7B. Each component is designed with a specific functionality, creating a modular architecture that couples symbolic reasoning with visual perception and enforces external verification, forming a unified and extensible reasoning pipeline. These components include three specialized tools and a verifier module:
\begin{itemize}
\item \textbf{Locate Tool:} Given an image, a target object description, and optionally a region of interest, this tool returns the bounding box of the specified object. This capability supports coarse-to-fine localization in cluttered or ambiguous scenes.
\item \textbf{VLM Understanding Tool:} This tool takes an image and a query as input, utilizing a powerful vision-language model (VLM) to perform semantic understanding and generate a natural language answer accompanied by its reasoning.
\item \textbf{Adjust Bbox Tool:} Given a current bounding box, a coordinate-guided reference image, and a textual instruction (e.g., "expand leftward", "shrink top edge"), this tool refines the bounding box accordingly.
\item \textbf{Verifier Module:} This module verifies whether the current output (be it an answer or a region) satisfies the query requirement. It also helps determine whether further zooming or refinement is necessary by providing critical feedback.
\end{itemize}

\begin{figure}[ht!]
    \centering
    \includegraphics[width=0.85\textwidth]{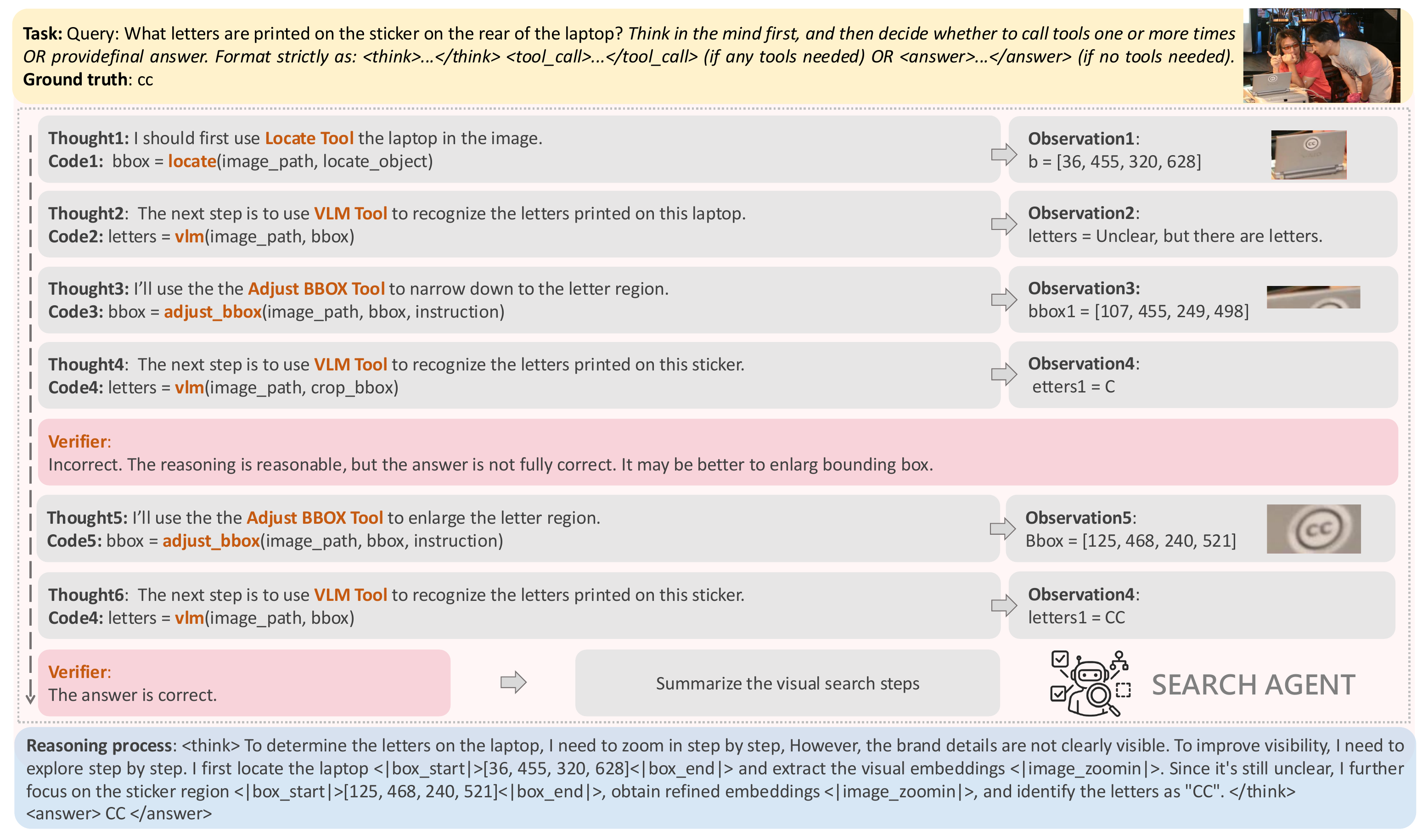}
      \caption{Illustration of the visual search agent that progressively locates regions necessary for answering the query. Sampled queries, images, and answers from the source datasets combined with the prompt are denoted in yellow. The reasoning steps of the agent are shown in pink. The summarized reasoning process is shown in blue.}
\label{fig:datapipeline}
\end{figure}

As illustrated in Fig.~\ref{fig:datapipeline}, our agent's reasoning and verification process is guided by a task prompt and in-context examples to iteratively find the correct visual search path. The process begins with a query and an image, while the ground-truth answer is provided exclusively to the verification module to guide exploration. The agent then engages in a step-by-step reasoning loop: it identifies and refines key bounding boxes before invoking an understanding tool to generate a grounded answer. The verifier module immediately evaluates the output; if correct, the trajectory is rewritten into a coherent explanation (the reasoning traces) by DeepSeek-V3. If incorrect, the module provides explicit feedback to guide the next round of exploration. This self-correction cycle, which continues until the correct answer is found or a step limit is reached, is crucial for generating faithful and interpretable reasoning traces.

Our data generation pipeline differs from conventional object detection, as the agent employs iterative reasoning with tools rather than direct detection to infer key regions. By dynamically adjusting bounding boxes based on multimodal tools' responses, the agent narrows from coarse localization to fine-grained grounding, focusing computation effectively. A verification module enhances reliability by checking answers against the ground truth, while contextual cues ensure a coherent reasoning trajectory.
Crucially, the resulting dataset explicitly records the reasoning steps involved in this grounding and adjustment process. This provides supervision for the intermediate process itself, not just the final answer, which is critical for training VLMs to develop interpretable, step-by-step reasoning. This approach enables the agent to ground small or ambiguous objects, such as text regions, beyond the reach of standard detectors by combining visual cues, tool operations, and commonsense reasoning.

\begin{figure}[ht]
    \centering
    \includegraphics[width=0.85\textwidth]{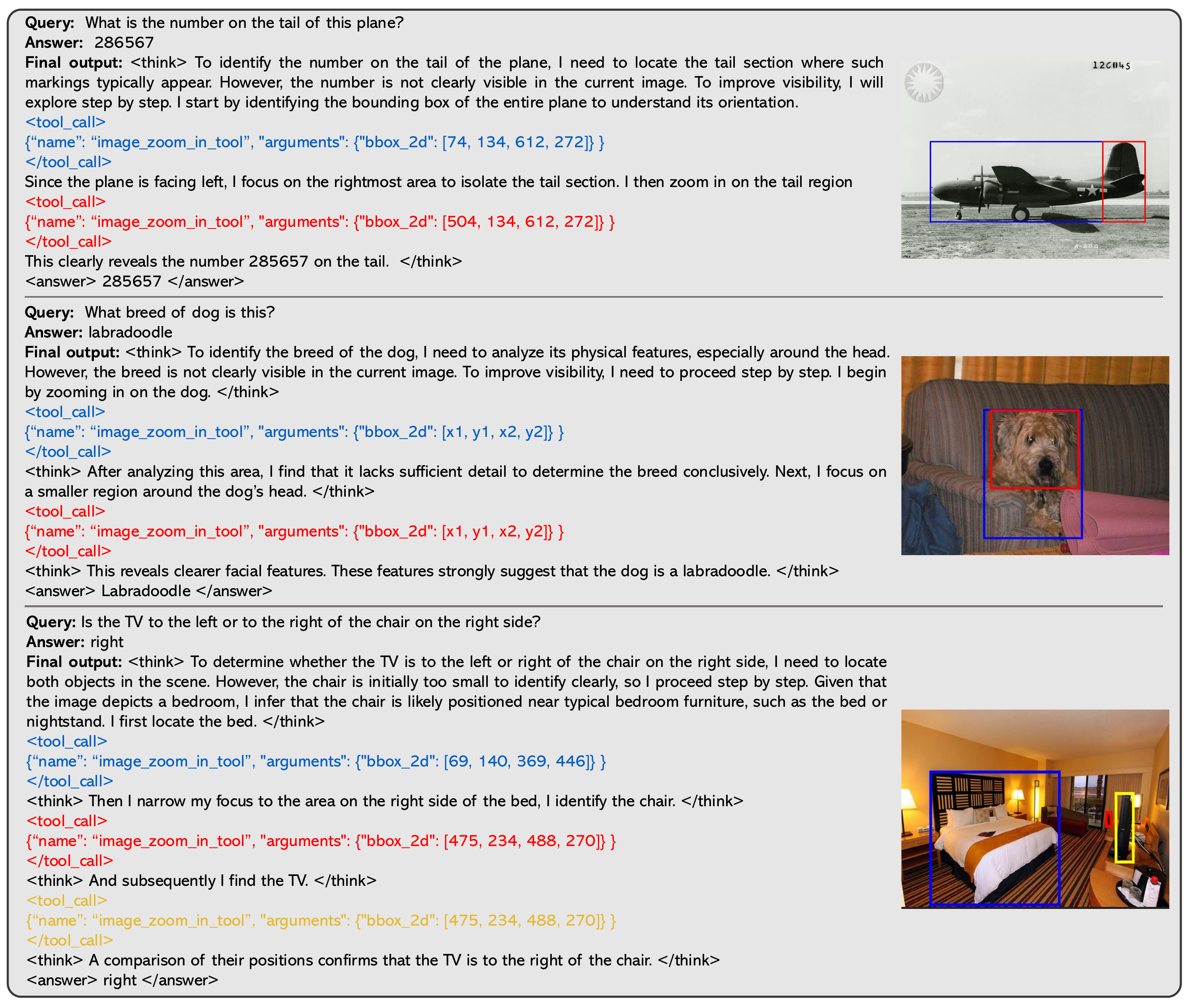}
      \caption{Examples of our generated Adaptive-CoF data via the visual search agent. Bounding boxes in red, blue, and yellow denote image regions focused in the first, second, and third steps, respectively. }
      \vskip -0.15in
\label{fig:generate_cases}
\end{figure}

As shown in \cref{fig:generate_cases}, our pipeline handles complex cases where standard detectors might fail. For instance, the agent identifies an airplane's tail using iterative reasoning, recognizes a dog's breed through trial-and-error zooming on its head, and locates a small chair by leveraging spatial reasoning about its proximity to a television. These examples highlight our agent's strength in grounding informative regions through a combination of iterative zoom-in tool use, spatial reasoning, and commonsense inference.

\begin{figure}
  \centering
  \subfigure[Data Number Distribution]{
  \includegraphics[width=0.28\textwidth]{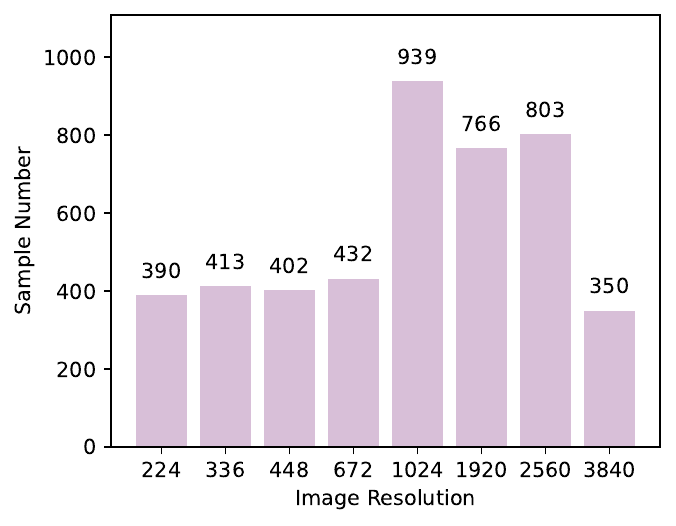}\label{fig:word_num}}
  \subfigure[Zooming in Distribution]{
  \includegraphics[width=0.28\textwidth]{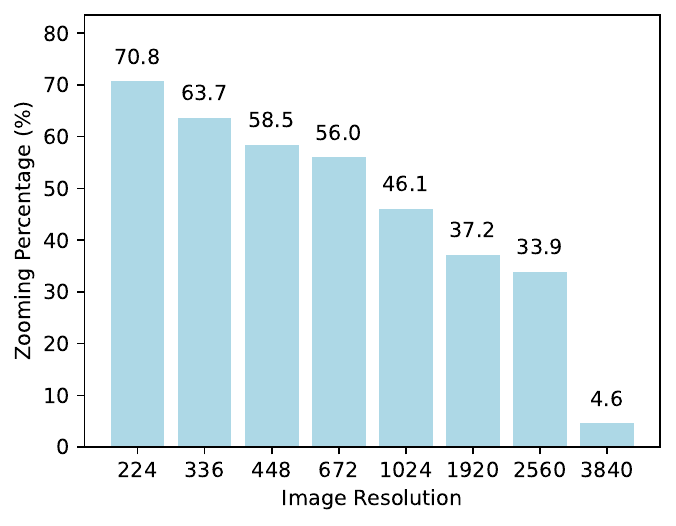}\label{fig:box_size_num1}}
  \subfigure[Reasoning Turn Distribution]{
  \includegraphics[width=0.28\textwidth]{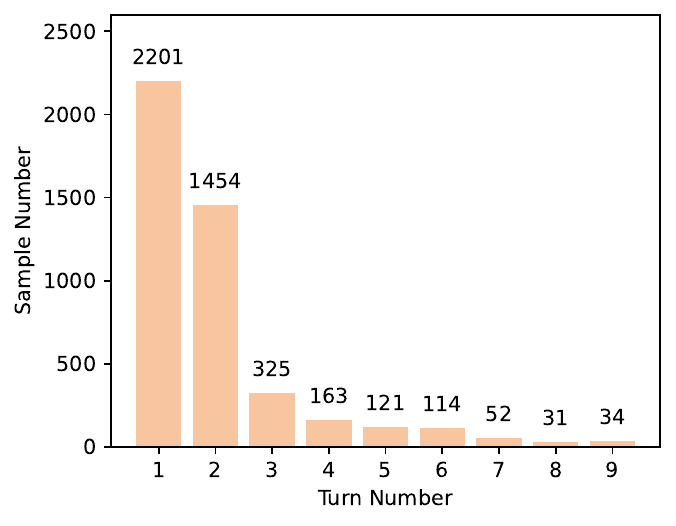}\label{fig:turn_num}}  
  \caption{Data statistics of MM-Adaptive-CoF SFT dataset.}
  \vskip -0.15in
  \label{fig:statistics}
\end{figure}

\subsubsection{Dataset Analysis}

We provide three key statistics to characterize the collected MM-Adaptive-CoF SFT data:  

\begin{itemize}


  \item \textbf{Sample Count.} 
    The number of samples across different image resolutions is shown in ~\cref{fig:word_num}. 
    The counts are generally comparable across resolutions, indicating a balanced distribution overall. 
    Samples with higher resolutions are slightly more frequent, as our task primarily focuses on small-object understanding in high-resolution images, and the original data also tend to have higher native resolutions.
    
    \item \textbf{Zoom-in Distribution.} 
    The proportion of samples requiring zoom-in operations is illustrated in ~\cref{fig:box_size_num1}. 
    Since our dataset is collected from the real reasoning process of a search agent, the need for zoom-in operations decreases as image resolution increases. 
    This aligns with intuition — higher-resolution images naturally provide richer visual details, thus reducing reliance on localized zooming during question answering.
    
    \item \textbf{Reasoning Turn Distribution.} 
    The distribution of reasoning turns is shown in ~\cref{fig:turn_num}. 
    The numbers of samples requiring a single reasoning turn and those requiring multiple turns are roughly comparable, averaging around 2.2k each. 
    As the number of reasoning turns increases, the sample frequency drops sharply, and extremely long reasoning trajectories involving many steps are exceedingly rare, indicating that such complex multi-step reasoning processes constitute only a small fraction of the dataset.

\end{itemize}

Overall, these statistics show that the dataset is balanced across resolutions, captures realistic zoom-in requirements, and covers diverse reasoning depths. Such characteristics encourage models to develop adaptive reasoning abilities, i.e., deciding when a single-step answer suffices and when multi-step reasoning with zoom-in operations is necessary, thereby improving both efficiency and interpretability in multimodal reasoning tasks.

\subsection{MM-Adaptive-CoF Reinforcement Learning (RL) Dataset}

The Reinforcement Learning (RL) stage leverages a specialized dataset comprising approximately 10,000 diverse query-response instances in total. Among these, 1,500 QA pairs are derived from the carefully constructed proprietary MM-Adaptive-CoF data. To enhance the diversity and robustness of the RL training, these proprietary QA pairs are augmented by merging them with several existing, publicly available datasets. Specifically, the RL dataset incorporates data from VisDrone~\cite{zhu2021detection}, ArxivQA~\cite{li2024multimodal}, and ThinkLite-VL~\cite{wang2025sota}. The inclusion of these external datasets broadens the domain coverage, introduces a wider array of visual complexities, and presents diverse reasoning challenges, thereby enabling the model to learn more generalized and robust adaptive reasoning capabilities through reinforcement learning.
We denote this reinforcement learning dataset as $\mathbb{U}$, which is used for policy optimization in the RL stage.

\section{Training}

We adopt a two-stage training pipeline consisting of supervised fine-tuning (SFT) followed by reinforcement learning (RL). The collected MM-Adaptive-CoF data are used to train Qwen2.5-VL-7B, which incorporates a ViT backbone as the visual encoder, a two-layer multilayer perceptron as the projector, and Qwen2.5-7B as the language model. 

\subsection{Cold-Start Tuning}
The cold-start stage aims to familiarize the model with the expected reasoning formats and establish a stable initialization for subsequent reinforcement learning. Specifically, the model is trained to handle both purely textual reasoning and multimodal reasoning that interleaves textual outputs with bounding box predictions and zoom-in operations. To achieve this, we freeze the ViT backbone and apply LoRA fine-tuning to the remaining components, including the multilayer perceptron projector and the language model. 
In this stage, the model is trained on the MM-Adaptive-CoF SFT dataset. 
Formally, given a training dataset $\mathbb{D}$, an $n$-step Adaptive-CoF instance is represented as $(I, q, O=\{r_1, o_1, t_1, \ldots, r_n, t_n\})$, where $I$ denotes the input image, $q$ the query, and $O$ the complete reasoning trajectory of $n$ steps. Each step consists of key regions $r_i$, the associated textual output $o_i$, and the visual embedding $t_i$, which is generated from $r_i$ during training. The model is fine-tuned with the standard cross-entropy loss objective, which enables it to generate coherent reasoning traces and prepares it for reward-based optimization in the RL stage,

\begin{equation}
\begin{aligned}
\min_{\theta} \mathbb{E}_{(I, q, O)\sim \mathbb{D} }\left[ - \sum_{i=1}^{n} \log \pi_{\theta} (r_i, o_i | I, q, \{r_j, o_j, t_j\}_{j=1}^{j=i-1} )  \right].
\end{aligned}
\end{equation}

\subsection{Reinforcement Learning}

In the RL stage, the model is further optimized with the Group Relative Policy Optimization (GRPO) algorithm~\cite{shao2024deepseekmath}.
The model is trained on the MM-Adaptive-CoF RL dataset in this stage.
Formally, given a training dataset $\mathbb{U}$, an instance is represented as $(I, q)$, where $I$ denotes the input image and $q$ the corresponding query. 
For each $(I, q) \sim \mathbb{U}$, GRPO samples a group of $G$ reasoning trajectories $\{O_1, O_2, \ldots, O_G\}$ from the old policy $\pi_{\text{old}}(\cdot \mid I, q)$.
The policy $\pi_\theta$ is updated by maximizing the following objective:


\begin{equation}
\begin{aligned}
    \mathcal{J}_{\text{GRPO}}(\theta) = 
    &\mathbb{E}_{(I, q)\sim \mathbb{U},\, \{O_i\}_{i=1}^G \sim \pi_{\text{old}}(\cdot \mid I, q)} \\
    &\bigg[ \frac{1}{G} \sum_{i=1}^G \left( \frac{1}{\sum_{t=1}^{|O_i|} \mathbbm{1}(O_{i,t})} \sum_{t=1}^{|O_i|} \mathbbm{1}(O_{i,t}) \cdot \min\left( p_{i,t} \hat{A}_{i,t},\ \text{clip}\left( p_{i,t}, 1 - \epsilon,\ 1 + \epsilon \right) \hat{A}_{i,t} \right) \right) \bigg] \\
    &- \beta\, \mathbb{D}_{\text{KL}}[\pi_\theta \,\|\, \pi_{\text{ref}}],
\end{aligned}
\label{eq:grpo_objective}
\end{equation}

where $p_{i,t} = \frac{\pi_\theta(O_{i,t} \mid I, q, O_{i,<t})}{\pi_{\text{old}}(O_{i,t} \mid I, q, O_{i,<t})}$ is the probability ratio for the $t$-th token $O_{i,t}$ in the $i$-th trajectory $O_i$. The advantage estimate $\hat{A}_{i,t}$, which is computed based on the relative ranking of rewards within the sampled group, is specifically given by the group-relative advantage ($A_{i}$) for the trajectory $O_i$, 
\begin{equation}
\hat{A}_{i,t} = \frac{r_i - \mu_r}{\sigma_r + \varepsilon},
\end{equation}
where $r_i$ is the reward, $\mu_r$ and $\sigma_r$ are the mean and standard deviation of the group's rewards, respectively, and $\varepsilon$ is a small constant for stability.
The indicator function $\mathbbm{1}(O_{i,t})$ selects valid reasoning tokens that should contribute to the loss. 
Specifically, it is defined as:
\begin{equation}
\mathbbm{1}(O_{i,t}) =
\begin{cases}
1, & \text{if } O_{i,t} \text{ is a text reasoning token},\\[4pt]
0, & \text{if } O_{i,t} \text{ is a vision token or padding token}.
\end{cases}
\end{equation}

This design ensures that the policy gradient is computed only on textual reasoning outputs, 
while excluding visual tokens or non-semantic placeholders from the optimization process.

\subsubsection{Adaptive Group-Aware Reward}

The reward signal serves as the optimization objective, directly guiding the policy model's behavior during training. To enable the model to achieve high accuracy while learning an adaptive strategy for improving efficiency, we introduce an adaptive group-aware reward (AGAR) instead of a simple binary reward, which adjusts rewards for complex reasoning based on the collective performance of the sampled group of $G$ rollouts. This design encourages the policy to be highly efficient when a task is simple, but to engage in deeper reasoning when necessary.

We first define the essential indicator variables for each rollout response $y_i$.
The correctness $c_i$ and format validity $f_i$ of the rollout are defined as binary variables:
\begin{equation}
c_i =
\begin{cases}
1, & \text{if the answer is correct} \\
0, & \text{otherwise}
\end{cases} ,
\quad
f_i =
\begin{cases}
1, & \text{if the response format is valid} \\
0, & \text{otherwise}
\end{cases}.
\end{equation}

Then, we define a group-level signal $g$ as an indicator that equals 1 if any correct $\texttt{direct}$ answer exists within the group, and 0 otherwise. 
This signal $g$ determines the policy's reward discount factor for inefficient response.
Two indicators $d_i$ and $z_i$ are used to distinguish the two reasoning strategies of the response $y_i$: $d_i = \mathbbm{1}_{\{\mathrm{format}(y_i)=\text{\texttt{direct}}\}}$ for a \texttt{direct} answer and $z_i = \mathbbm{1}_{\{\mathrm{format}(y_i)=\text{\texttt{zoomin}}\}}$ for a reasoning path that involves $\texttt{zoom-in}$ steps.


\medskip

The final reward $r_i$ for a rollout response $y_i$ is calculated by combining the correctness and format components. The weight for the $\texttt{zoom-in}$ path is dynamically adjusted by the group-level signal $g$:
\begin{equation}
    r_i = c_i\big(d_i \cdot 1 + z_i \cdot (1 - \delta \cdot g)\big) + (1 - c_i)\cdot (\gamma \cdot f_i),
\end{equation}
where $c_i$ denotes correctness, $\delta=0.2$ is the penalty factor, and $\gamma=0.1$ is a small format bonus granted only when the answer is incorrect but the output format is valid.
This structure ensures that a correct $\texttt{direct}$ answer always receives the maximum reward (the value is 1), promoting efficiency. The $\texttt{zoom-in}$ path is rewarded with $(1-\delta)$ when there is an efficient solution exists ($g=1$), thereby promoting maximum efficiency in the sampled group.

During the computation of the GRPO objective, we mask out the visual tokens ($t_j$),
which prevents these external tokens (visual embeddings) from contributing to the SFT and RL losses, ensuring training stability and preserving the model’s inherent reasoning sequences from disruption. The policy is primarily updated based on its ability to generate the \emph{decision} (region $a_i$) and the \emph{explanation} ($o_i$).

\section{Experiments}


\subsection{Experimental Setting}

\subsubsection{Datasets and Benchmarks}
To thoroughly assess our model, we strategically selected benchmarks designed to collectively evaluate three critical aspects: its proficiency in high-resolution image perception, its comprehensive multimodal reasoning, and its robustness against hallucinations.

\textbf{High-Resolution Fine-Grained QA.} This task focuses on the primary challenge of perceiving and reasoning about details in high-resolution images. We evaluate performance on this task using benchmarks such as $V^{\star}$ Bench~\cite{wu2024v}, which measures fine-grained attribute recognition and spatial reasoning, and HR-Bench~\cite{wang2024divide}, for visual query answering on 4K and 8K images.

\textbf{Comprehensive Reasoning and VQA.} To evaluate a wider set of reasoning skills, we selected a suite of benchmarks that test diverse capabilities on real-world images. This suite includes the MME-RealWorld~\cite{zhang2024mmerealworld} benchmark, which provides a multifaceted evaluation of both perception (e.g., OCR, detection) and reasoning (e.g., logic, math). Furthermore, we incorporated SpatialScore~\cite{wu2025spatialscore} to specifically assess complex spatial reasoning, and TallyQA~\cite{acharya2019tallyqa} to evaluate numerical and counting skills. The main goal of this suite is to test a broad range of capabilities beyond just high-resolution perception.

\textbf{Hallucination Benchmark.} To measure factual grounding and reliability, we use the POPE~\cite{ma2022pope} benchmark. This benchmark evaluates the model's tendency to hallucinate objects, requiring that its responses be faithfully aligned with the visual content.


\subsubsection{Implementation Details}

The model is trained in two distinct stages using Qwen2.5-VL-7B~\cite{bai2025qwen2} as the base: supervised fine-tuning followed by reinforcement learning. The specific hyperparameters and configurations for each stage are detailed in Table~\ref{tab:training_details}.

\begin{table}[htbp]
\centering
\caption{Training details for the Supervised Fine-tuning (SFT) and Reinforcement Learning (RL) stages.}\vspace{1pt}
\label{tab:training_details}
\resizebox{0.68\textwidth}{!}{
\begin{tabular}{@{}lcc@{}}
\toprule
\textbf{Configuration} & \textbf{Supervised Fine-tuning (SFT)} & \textbf{Reinforcement Learning (RL)} \\ 
\midrule
Tuned Components      & Projector \& LLM              & Projector \& LLM                   \\ 
LoRA Rank             & 32                            & N/A                                \\ 
\addlinespace 
Optimizer             & AdamW                         & AdamW                              \\ 
Learning Rate         & 1e-5                          & 1e-6                               \\
Training Epochs       & 3                             & 2                                  \\ 
\addlinespace
Global Batch Size     & 16                            & 32                                  \\ 
Hardware              & 4 $\times$ NVIDIA A100        & 8 $\times$ NVIDIA A100             \\
Training Time (approx.) & $\sim$6 hours                 & $\sim$32 hours                     \\ 
\bottomrule
\end{tabular}
}
\end{table}

\subsubsection{Baselines}
We compare our Adaptive-CoF model against two distinct categories of state-of-the-art vision-language models.

\textbf{General Models.}
This group includes leading closed-source models like GPT-4o~\cite{hurst2024gpt}, o3~\cite{openai2025o3}, and Gemini~\cite{gemini2025pushing}, as well as powerful open-source models like LLaVA-OneVision~\cite{an2025llava}. Crucially, it also includes our base model, Qwen2.5-VL~\cite{bai2025qwen2}, which serves as the primary point of comparison to directly measure the impact of our Adaptive-CoF methodology.

\textbf{Visual Grounded Reasoning Models.}
This category comprises models that embody the "think with image" paradigm, actively integrating new visual information into their reasoning process. We divide these models into two main architectural approaches.
The first are multi-stage frameworks, such as SEAL~\cite{wu2024v}, DyFo~\cite{li2025dyfo}, Visual Sketchpad (GPT-4o)\cite{hu2024visual}, IVM-Enhance (GPT-4V)\cite{zheng2024instruction}, and PaLI-3-VPD~\cite{hu2024visualpali}. These methods explore visual content or extract detailed information using a pipeline of separate steps. For example, they might first use an object detector and then pass the results to a different model for reasoning. These components are often not trained together as a single system.
The second are end-to-end models, including recent works like DeepEyes~\cite{zheng2025deepeyes} and Pixel Reasoner~\cite{su2025pixel}. These models are methodologically the closest to our own, as they use a single, unified model that learns to generate zoom-in tool calls as part of its internal reasoning process. A critical distinction, however, is that these models are typically encouraged to proactively zoom in to explore visual details. In contrast, our Adaptive-CoF is designed to invoke exploration adaptively, only when it is deemed necessary for the task at hand.

\subsection{Main Result}
\subsubsection{Results on General Models}


As detailed in Table~\ref{tab:highres_gm}, Table~\ref{tab:mme_gm}, and Table~\ref{tab:pope_gm_spatial}, Adaptive-CoF demonstrates a ignificant improvement in performance compared to general-purpose vision-language models. Our model not only achieves substantial gains over its baseline but also establishes itself as a state-of-the-art (SOTA) open-source model in the 7B-8B parameter class across a wide range of benchmarks.

On high-resolution benchmarks (Table~\ref{tab:highres_gm}), the model consistently outperforms other leading open-source models like LLaVA-OneVision and InternVL3. Most notably, it improves upon its Qwen2.5-VL-7B base by a remarkable 18.9\% on the $V^\star$ Bench.
This dominance extends to comprehensive reasoning, even against significantly larger models. On the MME-RealWorld-Lite benchmark (Table~\ref{tab:mme_gm}), Adaptive-CoF achieves an overall score of 50.9, the highest among all listed models, including the 72B variants of LLaVA-OneVision and Qwen2.5-VL, driven by its exceptional ability in fine-grained perception and reasoning.
Furthermore, Adaptive-CoF demonstrates superior reliability and specialized reasoning (Table~\ref{tab:pope_gm_spatial}). It sets a new SOTA on the POPE benchmark for hallucination reduction with a 3.4\% gain over its base model. In quantitative reasoning, its score of 75.0 on TallyQA is highly competitive, marking a significant 6.4\% improvement. Crucially, on the SpatialScore benchmark, it also secures a leading score of 20.6, representing a notable 5.4\% gain. These results underscore its position as a leading model in robust perception and reasoning.

\begin{table}[ht!]
\centering
\caption{Performance comparison against general models on high-resolution benchmarks ($V^{\star}$ Bench, HR-Bench). E2E denotes end-to-end models. The best performance is highlighted in \textbf{bold}, and the second-best performances are highlighted in \underline{underline}.}\vspace{1pt}
\label{tab:highres_gm}
\resizebox{0.80\textwidth}{!}{
\begin{tabular}{l c c | c c c | c c c | c c c}
\toprule
\textbf{Model} & \textbf{E2E} & \textbf{Param Size} 
& \multicolumn{3}{c|}{\textbf{$V^{\star}$ Bench}} 
& \multicolumn{3}{c|}{\textbf{HR-Bench 4K}} 
& \multicolumn{3}{c}{\textbf{HR-Bench 8K}} \\
\cmidrule[0.2pt](lr){4-6} \cmidrule[0.2pt](lr){7-9} \cmidrule[0.2pt](lr){10-12}
& & 
& Attr & Spatial & Overall 
& FSP & FCP & Overall 
& FSP & FCP & Overall \\
\midrule
\multicolumn{12}{c}{\textbf{Private Models}} \\
\midrule
GPT-4o\cite{hurst2024gpt} & \checkmark & -- & -- & -- & 66.0 & 70.0 & 48.0 & 59.0 & 62.0 & 49.0 & 55.5 \\
o3\cite{openai2025o3} & \checkmark & -- & -- & -- & \textbf{95.7} & -- & -- & -- & -- & -- & -- \\
Gemini-2.0-Flash\cite{gemini2025pushing} & \checkmark & -- & -- & -- & 73.2 & -- & -- & -- & -- & -- & -- \\
Gemini-2.5-Pro\cite{gemini2025pushing} & \checkmark & -- & -- & -- & 79.2 & -- & -- & -- & -- & -- & -- \\
\midrule
\multicolumn{12}{c}{\textbf{Open-source General Models}} \\
\midrule
LLaVA-OneVision\cite{an2025llava} & \checkmark & 7B & \uline{75.7} & \uline{75.0} & 75.4 & 72.0 & 54.0 & 63.0 & 67.3 & \uline{52.3} & 59.8 \\
InternVL3\cite{chen2024internvl3} & \checkmark & 8B & 73.0 & 71.1 & 72.3 & 79.3 & \textbf{62.3} & \uline{70.8} & 64.3 & \textbf{59.8} & 62.0 \\
Qwen2.5-VL\cite{bai2025qwen2} & \checkmark & 7B & 73.9 & 67.1 & 71.2 & \uline{85.2} & 52.2 & 68.8 & \uline{78.8} & 51.8 & \uline{65.3} \\
\midrule
\rowcolor{rowgray}
\textbf{Adaptive-CoF} & \checkmark & 7B & \textbf{92.2} & \textbf{86.4} & \uline{90.1} 
& \textbf{88.3} & \uline{58.8} & \textbf{73.5} 
& \textbf{85.3} & 50.0 & \textbf{67.6} \\
\rowcolor{rowgray}
\textit{$\Delta$ (v.s. Qwen2.5-VL-7B)}
& -- & -- 
& \pos{18.3} & \pos{19.3} & \pos{18.9}
& \pos{3.1} & \pos{6.6} & \pos{4.7}
& \pos{6.5} & \negnum{1.8} & \pos{2.3} \\
\bottomrule
\end{tabular}
}
\end{table}

\begin{table}[ht!]
\centering
\caption{Performance comparison against general models on the MME-RealWorld-Lite benchmark.}\vspace{1pt}
\label{tab:mme_gm}
\resizebox{0.80\textwidth}{!}{
\begin{tabular}{l c c c c c c c c c c c}
\toprule
\textbf{Model} & \textbf{Param Size} &
\multicolumn{5}{c}{\textbf{Perception}} &
\multicolumn{4}{c}{\textbf{Reasoning}} &
\textbf{Overall} \\
\cmidrule[0.2pt](lr){3-7} \cmidrule[0.2pt](lr){8-11}
&  & OCR & RS & DT & MO & AD & OCR & DT & MO & AD & \\
\midrule
\multicolumn{12}{c}{\textbf{Private Models}} \\
\midrule
GPT-4o~\cite{hurst2024gpt}
& -- & -- & -- & -- & -- & -- & -- & -- & -- & -- & 45.2 \\
\midrule
\multicolumn{12}{c}{\textbf{Open-source General Models}} \\
\midrule
InternVL3~\cite{chen2024internvl3}
& 8B & 83.6 & 49.3 & 75.0 & 34.5 & 36.9 & 70.0 & 44.0 & \uline{40.0} & \uline{37.0} & 47.9 \\
LLaVA-OneVision~\cite{an2025llava}
& 7B & 80.0 & 40.0 & 56.0 & 31.7 & 39.4 & 65.0 & 33.0 & 38.0 & 32.0 & 43.7 \\
LLaVA-OneVision~\cite{an2025llava}
& 72B & 79.2 & \uline{50.7} & 67.0 & \uline{37.9} & \uline{40.0} & \textbf{76.0} & 41.0 &38.7 & \textbf{39.3} & \uline{48.7} \\
Qwen2.5-VL~\cite{bai2025qwen2}
& 7B & 87.6 & 32.7 & 83.0 & 27.3 & 30.0 & 72.0 & \uline{62.0} & 28.7 & 23.0 & 42.3 \\
Qwen2.5-VL~\cite{bai2025qwen2}
& 32B & 87.2 & 40.7 & 83.0 & 29.5 & \textbf{40.7} & 74.0 & 60.0 & 27.3 & 29.5 & 45.6 \\
Qwen2.5-VL~\cite{bai2025qwen2}
& 72B & 90.8 & 34.0 & \textbf{87.0} & 27.9 & 30.6 & \uline{74.0} & 61.0 & 26.7 & 25.5 & 43.7 \\
\midrule
\rowcolor{rowgray}
\textbf{Adaptive-CoF}
& 7B & \uline{90.0} & \textbf{55.3} & \uline{83.0} & \textbf{42.6} & 35.7 & 68.0 & \textbf{64.0} & \textbf{48.7} & 31.3 & \textbf{50.9} \\
\rowcolor{rowgray}
\textit{$\Delta$ (vs. Qwen2.5-VL-7B)}
& -- & \pos{2.4} & \pos{22.6} & \negnum{0.0} & \pos{15.3} & \pos{5.7}
& \negnum{4.0} & \pos{2.0} & \pos{20.0} & \pos{8.3}
& \pos{8.6} \\
\bottomrule
\end{tabular}
}
\end{table}

\begin{table}[ht!]
\centering
\caption{Performance comparison against general models on hallucination (POPE), VQA (TallyQA), and spatial reasoning benchmarks(SpatialScore).}\vspace{1pt}
\label{tab:pope_gm_spatial}
\resizebox{0.80\textwidth}{!}{ 
\begin{tabular}{l c c c c c c c} 
\toprule
\textbf{Model} & \textbf{Param Size} &
\multicolumn{4}{c}{\textbf{POPE}} &
\textbf{TallyQA} & \textbf{SpatialScore} \\ 
\cmidrule[0.2pt](lr){3-6}
&  & Adversarial & Popular & Random & Overall \\
\midrule
\multicolumn{8}{c}{\textbf{Private Models}} \\
\midrule
GPT-4o~\cite{hurst2024gpt}
& -- & -- & -- & -- & -- & 73.0 & \textbf{30.6} \\
Gemini-2.0-Flash~\cite{gemini2025pushing}
& -- & -- & -- & -- & -- & 73.8 & -- \\
Gemini-2.5-Pro~\cite{gemini2025pushing}
& -- & -- & -- & -- & -- & \uline{74.0} & -- \\
\midrule
\multicolumn{8}{c}{\textbf{Open-source General Models}} \\
\midrule
LLaVA-OneVision~\cite{an2025llava}
& 7B & -- & -- & -- & \uline{88.4} & -- & -- \\
Qwen2.5-VL~\cite{bai2025qwen2}
& 7B & \textbf{85.9} & \uline{86.5} & \uline{87.2} & 85.9 & 68.6 & 15.2 \\
Qwen2.5-VL~\cite{bai2025qwen2}
& 72B & -- & -- & -- & 84.9 & -- & -- \\
\midrule
\rowcolor{rowgray}
\textbf{Adaptive-CoF}
& 7B & \uline{84.7} & \textbf{86.7} & \textbf{90.2} & \textbf{89.3} & \textbf{75.0} & \uline{20.6} \\
\rowcolor{rowgray}
\textit{$\Delta$ (v.s. Qwen2.5-VL-7B)}
& --
& \negnum{1.2} & \pos{0.2} & \pos{3.0} & \pos{3.4}
& \pos{6.4} & \pos{5.4} \\ 
\bottomrule
\end{tabular}
}
\end{table}

\subsubsection{Results on Visual Grounded Reasoning Models}


\begin{table}[ht!]
\centering
\caption{Performance comparison against visual grounded reasoning models on high-resolution benchmarks ($V^{\star}$ Bench, HR-Bench). E2E denotes end-to-end models.}\vspace{1pt}
\label{tab:highres_rm}
\resizebox{0.80\textwidth}{!}{
\begin{tabular}{l c c | c c c | c c c | c c c}
\toprule
\textbf{Model} & \textbf{E2E} & \textbf{Param Size} 
& \multicolumn{3}{c|}{\textbf{$V^{\star}$ Bench}} 
& \multicolumn{3}{c|}{\textbf{HR-Bench 4K}} 
& \multicolumn{3}{c}{\textbf{HR-Bench 8K}} \\
\cmidrule[0.2pt](lr){4-6} \cmidrule[0.2pt](lr){7-9} \cmidrule[0.2pt](lr){10-12}
& & 
& Attr & Spatial & Overall 
& FSP & FCP & Overall 
& FSP & FCP & Overall \\

\midrule
\multicolumn{12}{c}{\textbf{Visual Search / Zoom-in Frameworks (Non-E2E)}}\\
\midrule
Visual Sketchpad (GPT-4o)\cite{hu2024visual} & $\times$ & -- & -- & -- & 80.4 & -- & -- & -- & -- & -- & -- \\
IVM-Enhance (GPT-4V)\cite{zheng2024instruction} & $\times$ & -- & -- & -- & 81.2 & -- & -- & -- & -- & -- & -- \\
PaLI-3-VPD\cite{hu2024visualpali} & $\times$ & 7B & -- & -- & 70.9 & -- & -- & -- & -- & -- & -- \\
PaLI-3-VPD\cite{hu2024visualpali} & $\times$ & 55B & -- & -- & 76.6 & -- & -- & -- & -- & -- & -- \\
SEAL\cite{wu2024v} & $\times$ & 7B & 74.8 & 76.3 & 75.4 & -- & -- & -- & -- & -- & -- \\
DyFo\cite{li2025dyfo} & $\times$ & 7B & 80.0 & 82.9 & 81.2 & -- & -- & -- & -- & -- & -- \\
\midrule
\multicolumn{12}{c}{\textbf{Visual Grounded Reasoning Models}} \\
\midrule
ViGoRL\cite{Sarch2025GroundedRL} & \checkmark & 7B & -- & -- & 86.4 
& -- & -- & -- 
& -- & -- & -- \\
Pixel Reasoner\cite{su2025pixel} & \checkmark & 7B & -- & -- & 85.3 
& -- & -- & 71.9 
& -- & -- & 65.1 \\
DeepEyes\cite{zheng2025deepeyes} & \checkmark & 7B & 89.57 & \textbf{88.2} & \uline{89.0}
& \textbf{91.8} & \uline{54.8} & \uline{73.3}
& \uline{84.5} & \textbf{54.0} & \textbf{69.3} \\
\midrule
\rowcolor{rowgray}
\textbf{Adaptive-CoF} & \checkmark & 7B & \textbf{92.2} & \uline{86.4} & \textbf{90.1} 
& \uline{88.3} & \textbf{58.8} & \textbf{73.5} 
& \textbf{85.3} & 50.0 & \uline{67.6} \\
\bottomrule
\end{tabular}
}
\end{table}

\begin{table}[ht]
\centering
\caption{Performance comparison against end-to-end visual grounded reasoning models on the MME-RealWorld-Lite benchmark.}\vspace{1pt}
\label{tab:mme_rm}
\resizebox{0.80\textwidth}{!}{
\begin{tabular}{l c c c c c c c c c c c}
\toprule
\textbf{Model} & \textbf{Param Size} &
\multicolumn{5}{c}{\textbf{Perception}} &
\multicolumn{4}{c}{\textbf{Reasoning}} &
\textbf{Overall} \\
\cmidrule[0.2pt](lr){3-7} \cmidrule[0.2pt](lr){8-11}
&  & OCR & RS & DT & MO & AD & OCR & DT & MO & AD & \\
\midrule

Pixel-Reasoner~\cite{su2025pixel}
& 7B & 89.6 & 52.0 & \uline{86.0} & 38.9 & 30.9 & \uline{71.0} & \textbf{72.0} & \uline{46.0} & \uline{32.5} & 49.7 \\
DeepEyes~\cite{zheng2025deepeyes}
& 7B & 90.0 & \uline{52.7} & \textbf{89.0} & \uline{43.3} & \uline{33.4} & \textbf{76.0} & \uline{69.0} & 44.0 & \textbf{35.0} & \textbf{53.2} \\
\midrule
\rowcolor{rowgray}
\textbf{Adaptive-CoF}
& 7B & \textbf{90.0} & \textbf{55.3} & 83.0 & \uline{42.6} & \textbf{35.7} & 68.0 & 64.0 & \textbf{48.7} & 31.3 & \uline{50.9} \\
\bottomrule
\end{tabular}
}
\end{table}

\begin{table}[ht]
\centering
\caption{Comparison of average visual tokens processed by zoom-in operation and average zoom-in tool calls across various benchmarks. Lower values for tokens and zoom-in tool uses indicate greater efficiency.}\vspace{1pt}
\label{tab:efficiency_vs_deepeyes}
\resizebox{0.80\textwidth}{!}{
\begin{tabular}{l c | c c | c c | c c | c c}
\toprule
\textbf{Model} & \textbf{Param Size} & \multicolumn{2}{c|}{\textbf{$V^{\star}$ Bench}} & \multicolumn{2}{c|}{\textbf{HR-Bench 4K}} & \multicolumn{2}{c|}{\textbf{HR-Bench 8K}} & \multicolumn{2}{c}{\textbf{MME-RealWorld}} \\
\cmidrule(lr){3-4} \cmidrule(lr){5-6} \cmidrule(lr){7-8} \cmidrule(lr){9-10}
& & Tokens & Zoom-in & Tokens & Zoom-in & Tokens & Zoom-in & Tokens & Zoom-in \\
\midrule
Pixel Reasoner~\cite{su2025pixel}
& 7B & 408 & 0.8 & 575 & 0.9 & 601 & 0.9 & 456 & 0.8 \\ 
DeepEyes~\cite{zheng2025deepeyes}
& 7B & 75 & 2.0 & 930 & 1.8 & 710 & 1.7 & 426 & 1.8 \\ 
\midrule
\rowcolor{rowgray}
\textbf{Adaptive-CoF}
& 7B & 34 & 0.5 & 50 & 0.4 & 81 & 0.5 & 87 & 1.0 \\ 
\bottomrule
\end{tabular}
}
\end{table}

\begin{table}[ht]
\centering
\caption{Performance comparison against end-to-end visual grounded reasoning models on Hallucination (POPE), VQA (TallyQA), and Spatial Reasoning benchmarks(SpatialScore).}\vspace{1pt}
\label{tab:pope_rm_spatial}
\resizebox{0.75\textwidth}{!}{ 
\begin{tabular}{l c c c c c c c} 
\toprule
\textbf{Model} & \textbf{Param Size} &
\multicolumn{4}{c}{\textbf{POPE}} &
\textbf{TallyQA} & \textbf{SpatialScore} \\ 
\cmidrule[0.2pt](lr){3-6}
&  & Adversarial & Popular & Random & Overall \\
\midrule
ViGoRL\cite{Sarch2025GroundedRL}
& 7B & -- & -- & -- & 88.3 & -- & 19.5 \\
Pixel-Reasoner~\cite{su2025pixel}
& 7B & -- & -- & -- & -- & 73.8 & 20.2 \\
DeepEyes~\cite{zheng2025deepeyes}
& 7B & \uline{84.0} & \textbf{87.5} & \textbf{91.8} & \uline{87.7} & \textbf{76.8} & \uline{20.3} \\
\midrule
\rowcolor{rowgray}
\textbf{Adaptive-CoF}
& 7B & \textbf{84.7} & \uline{86.7} & \uline{90.2} & \textbf{89.3} & \uline{75.0} & \textbf{20.6} \\
\bottomrule
\end{tabular}
}
\end{table}

As detailed in Table~\ref{tab:highres_rm}, Table~\ref{tab:mme_rm}, Table~\ref{tab:efficiency_vs_deepeyes}, and Table~\ref{tab:pope_rm_spatial}, our Adaptive-CoF model demonstrates state-of-the-art (SOTA) performance and exceptional computational efficiency when compared against other visual grounded reasoning models.

On high-resolution fine-grained benchmarks (Table~\ref{tab:highres_rm}), our model establishes its leading position. It achieves a SOTA score of 90.1 on $V^{\star}$ Bench and secures the overall SOTA on the HR-Bench 4K benchmark, while remaining highly competitive on the more demanding HR-Bench 8K. In the comprehensive reasoning category (Table~\ref{tab:mme_rm}), Adaptive-CoF showcases robust generalization, achieving a score of 50.9 on the MME-RealWorld-Lite benchmark that is very close to the SOTA and leads in several key sub-categories.

Most notably, these top-tier results are achieved with remarkable efficiency (Table~\ref{tab:efficiency_vs_deepeyes}). A detailed comparison reveals that, compared to DeepEyes on HR-Bench 4K, Adaptive-CoF uses only about 5.4\% of the visual tokens and reduces zoom-in by 77.8\%; on $V^{\star}$ Bench, zoom-in operation is also reduced by 75\%.
This contrast further highlights how our adaptive strategy maintains comparable and even SOTA performance while drastically saving computational resources.
The table data also shows that while Pixel Reasoner invokes zoom-in operation less frequently than DeepEyes, it processes more visual tokens. This is because the \texttt{min pixels} parameter in its implementation is set to a large value, causing more tokens to be processed through upsampling.
Furthermore, Adaptive-CoF excels on specialized and reliability benchmarks (Table~\ref{tab:pope_rm_spatial}). It sets a new SOTA record on SpatialScore and achieves a highly competitive score on TallyQA for advanced spatial and quantitative reasoning. It also achieves a top score of 89.3 on the POPE benchmark for its low hallucination rate. In summary, Adaptive-CoF combines leading performance on diverse reasoning tasks with a massive leap in computational efficiency via its adaptive exploration strategy, setting a new standard for high-performance VLMs.

\subsection{Adaptive Zoom-in Analysis}

\subsubsection{Statistics and Results}

\begin{table*}[ht]
\centering
\caption{Comprehensive analysis of performance and efficiency on the $V^{\star}$ benchmark. The table details accuracy, computational cost (visual tokens, average zoom-in tool calls), and the resulting MITE Score. The analysis highlights the superior efficiency of our adaptive model, especially at high resolutions.}\vspace{1pt}
\label{tab:full_analysis_with_delta_acc}
\resizebox{0.75\textwidth}{!}{ 
\begin{tabular}{l c c c c c c c}
\toprule
\textbf{Model} & \textbf{224} & \textbf{336} & \textbf{448} & \textbf{672} & \textbf{1024} & \textbf{1920} & \textbf{2560} \\
\midrule
\multicolumn{8}{l}{\textit{\textbf{Accuracy (\%)}}} \\
Qwen2.5-VL-7B & 37.7 & 41.36 & 52.88 & \textbf{62.3} & 69.11 & 79.58 & 79.06 \\
\rowcolor{rowgray}
\textbf{Adaptive-CoF} & \textbf{39.27} & \textbf{43.98} & \textbf{53.93} & 60.21 & \textbf{75.39} & \textbf{87.96} & \textbf{89.53} \\
\rowcolor{rowgray}
\textit{$\Delta$ Acc (v.s. Qwen2.5-VL-7B)} & \pos{1.57} & \pos{2.62} & \pos{1.05} & \negnum{2.09} & \pos{6.28} & \pos{8.38} & \pos{10.47} \\
\midrule
\multicolumn{8}{l}{\textit{\textbf{Visual Tokens \& Zoom-in Tool Calls}}} \\
Qwen2.5-VL-7B (Tokens) & 54 & 127 & 244 & 549 & 1282 & 4307 & 4455 \\
\rowcolor{rowgray}
\textbf{Adaptive-CoF} (Tokens) & 107 & 176 & 280 & 590 & 1303 & 4340 & 4489 \\
\rowcolor{rowgray}
\textbf{Adaptive-CoF} (Zoom-in Tool Calls) & 2.5 & 2.05 & 1.65 & 1.03 & 0.75 & 0.52 & 0.53 \\
\midrule
\multicolumn{8}{l}{\textit{\textbf{MITE Score (Acc Gain per 100 Tokens)}}} \\
\rowcolor{rowgray}
\textbf{Adaptive-CoF (v.s. Qwen2.5-VL-7B)} & 3.0 & 5.4 & 2.9 & -5.1 & 29.9 & 25.4 & 30.8 \\
\bottomrule
\end{tabular}
}
\end{table*}

\begin{figure}[ht!]
  \centering
  \subfigure[Mean zoom-in operations across input resolutions]{
    \includegraphics[width=0.40\textwidth]{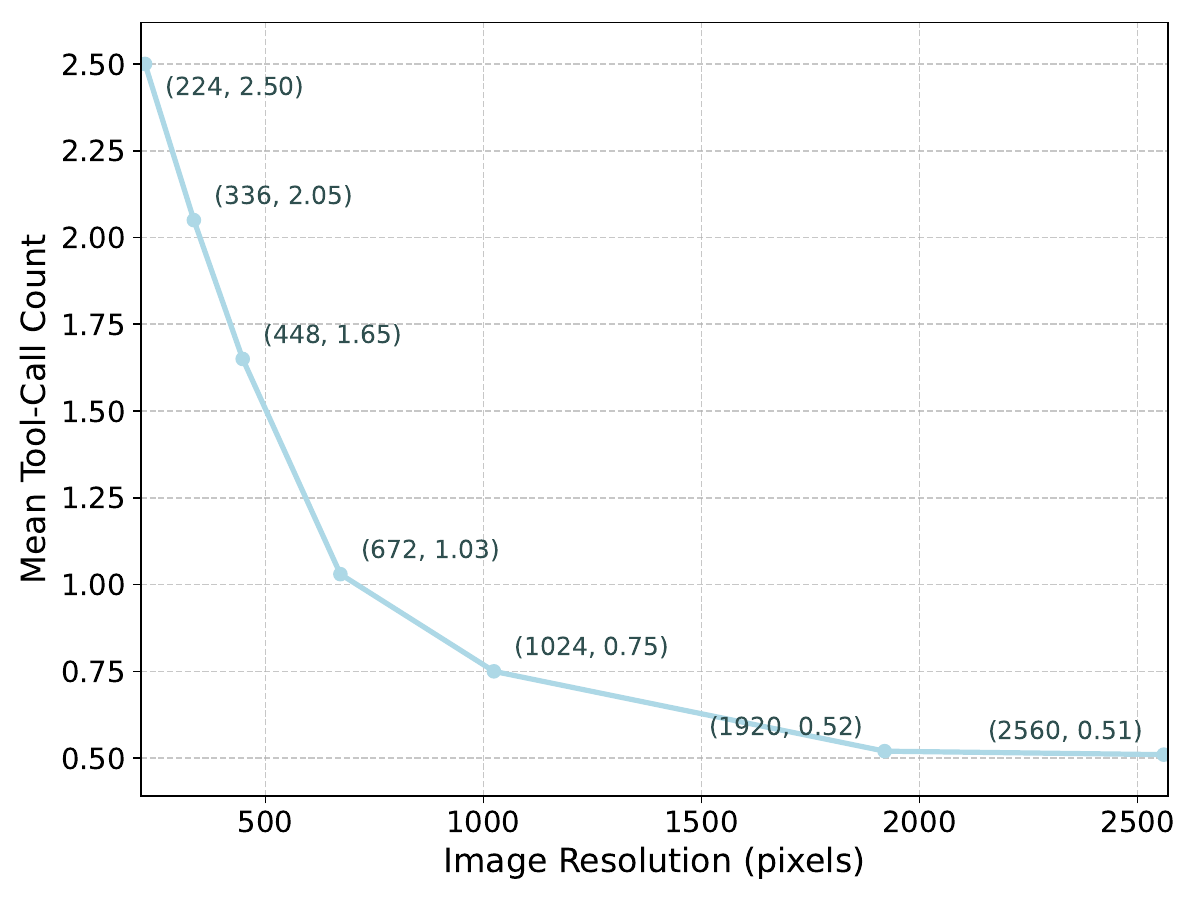}
    \label{fig:tool_call_mean}
  }
  \subfigure[Accuracy vs. visual tokens across resolutions]{
    \includegraphics[width=0.40\textwidth]{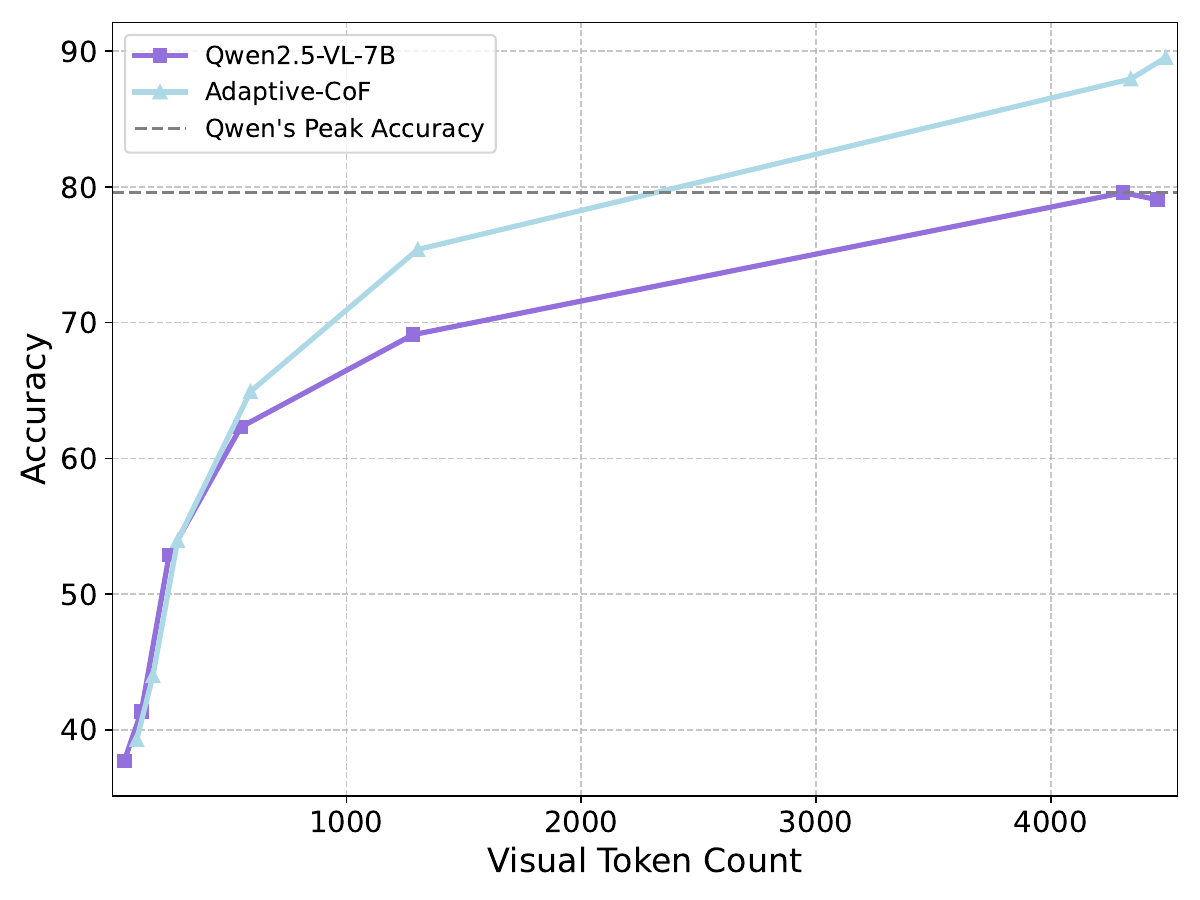}
    \label{fig:adaptive_performance_analysis}
  }
  \vskip -0.1in
  \caption{Analysis of the Adaptive-CoF model's efficiency and adaptive behavior.}
  \vskip -0.1in
  \label{fig:efficiency_analysis}
\end{figure}

A core strength of our Adaptive Chain-of-Focus (Adaptive-CoF) model is its ability to adaptively decide when to zoom in. To analyze this, we evaluated its performance and zoom-in frequency patterns across a wide range of image resolutions.
As shown in Figure~\ref{fig:tool_call_mean}, the model exhibits highly adaptive behavior. The average number of zoom-in tool calls is inversely correlated with image resolution. At lower resolutions (e.g., 224px), the model performs around 2.5 zoom-in operations on average to capture finer details, whereas at higher resolutions (e.g., 2560px), it makes only about 0.5 calls, correctly recognizing that sufficient visual detail is already available in the original image.
This intelligent allocation of resources leads to improved performance across the resolution spectrum, as detailed in the `Accuracy (\%)` section of Table~\ref{tab:full_analysis_with_delta_acc}. At low resolutions, more frequent zoom-ins give Adaptive-CoF a clear advantage, reflected by the positive gains in the `$\Delta$ Acc` row. Conversely, at very high resolutions (1920, 2560px), it excels by making fewer, more effective zoom-in operations, allowing it to focus on critical regions and avoid distraction. This results in a substantial performance increase, culminating in a remarkable 10.47\% gain at 2560px resolution, as shown in the final column of the `$\Delta$ Acc` row. This analysis confirms that Adaptive-CoF has learned an efficient and effective adaptive strategy: it zooms in when necessary to overcome low-resolution limitations and refrains when high-resolution input is sufficient.

\subsubsection{Efficiency Analysis}

A key advantage of our Adaptive-CoF model lies not only in its superior accuracy but also in its remarkable computational efficiency. To formally quantify this, we introduce the marginal improvement in token efficiency (MITE) score, defined as the accuracy gain per 100 additional visual tokens relative to the baseline:
\begin{equation}
\label{eq:mite}
\text{MITE} = \frac{\Delta \text{Accuracy}}{\Delta \text{Tokens}} \times 100.
\end{equation}
A higher MITE score signifies a more effective conversion of computational cost into performance gains. As detailed in Table~\ref{tab:full_analysis_with_delta_acc}, the MITE score confirms the benefit of our adaptive approach at high resolutions. While the score is modest at lower resolutions, it becomes exceptionally high for images above 1024px, peaking at 30.8. This indicates that our model's accuracy improvements significantly exceed its marginal token cost precisely where high-resolution reasoning is most critical.
This efficiency is most practically illustrated by comparing the total computational cost to reach peak performance, as shown in Figure~\ref{fig:adaptive_performance_analysis}. The baseline, Qwen2.5-VL-7B, reaches its peak accuracy of approximately 80\% only after processing over 4,000 visual tokens. In stark contrast, our Adaptive-CoF model surpasses this performance threshold while consuming just over 2,000 visual tokens. This represents a substantial computational saving of nearly 50\% to achieve the same level of peak performance as the baseline. This powerful combination of higher accuracy and reduced computational overhead underscores the effectiveness of our adaptive approach.

\begin{figure}[ht!]
    \centering
    \includegraphics[width=0.75\textwidth]{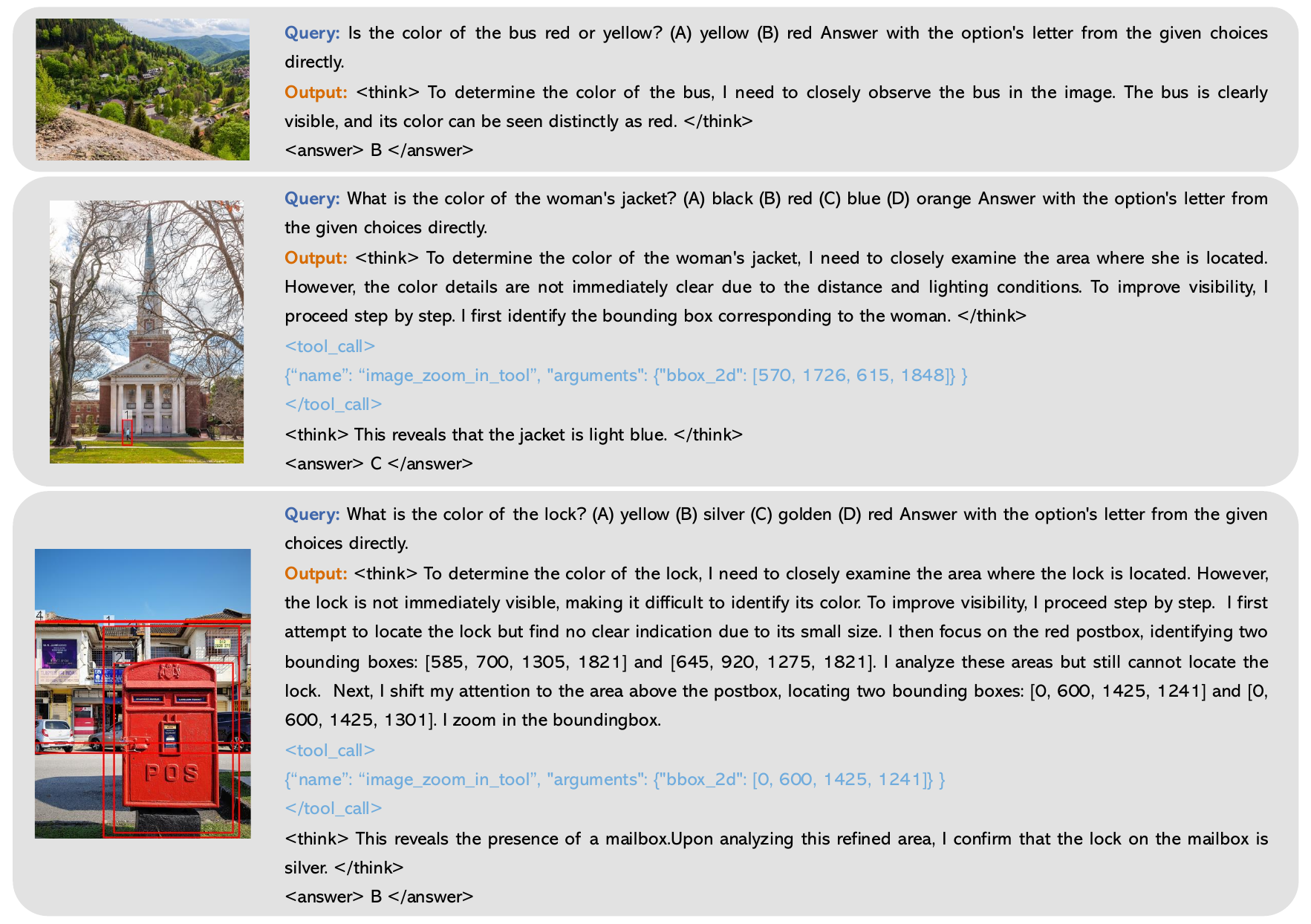}
  \vskip -0.1in
      \caption{
    Case studies demonstrating Adaptive-CoF's ability to adapt its reasoning strategy. 
    It handles simple tasks directly (top), uses a single zoom for moderately challenging tasks (middle), and engages in complex, iterative visual search for difficult, fine-grained queries (bottom).
}
      \vskip -0.1in
\label{fig:visualization}
\end{figure}

\begin{figure}[ht!]
    \centering
    \includegraphics[width=0.75\textwidth]{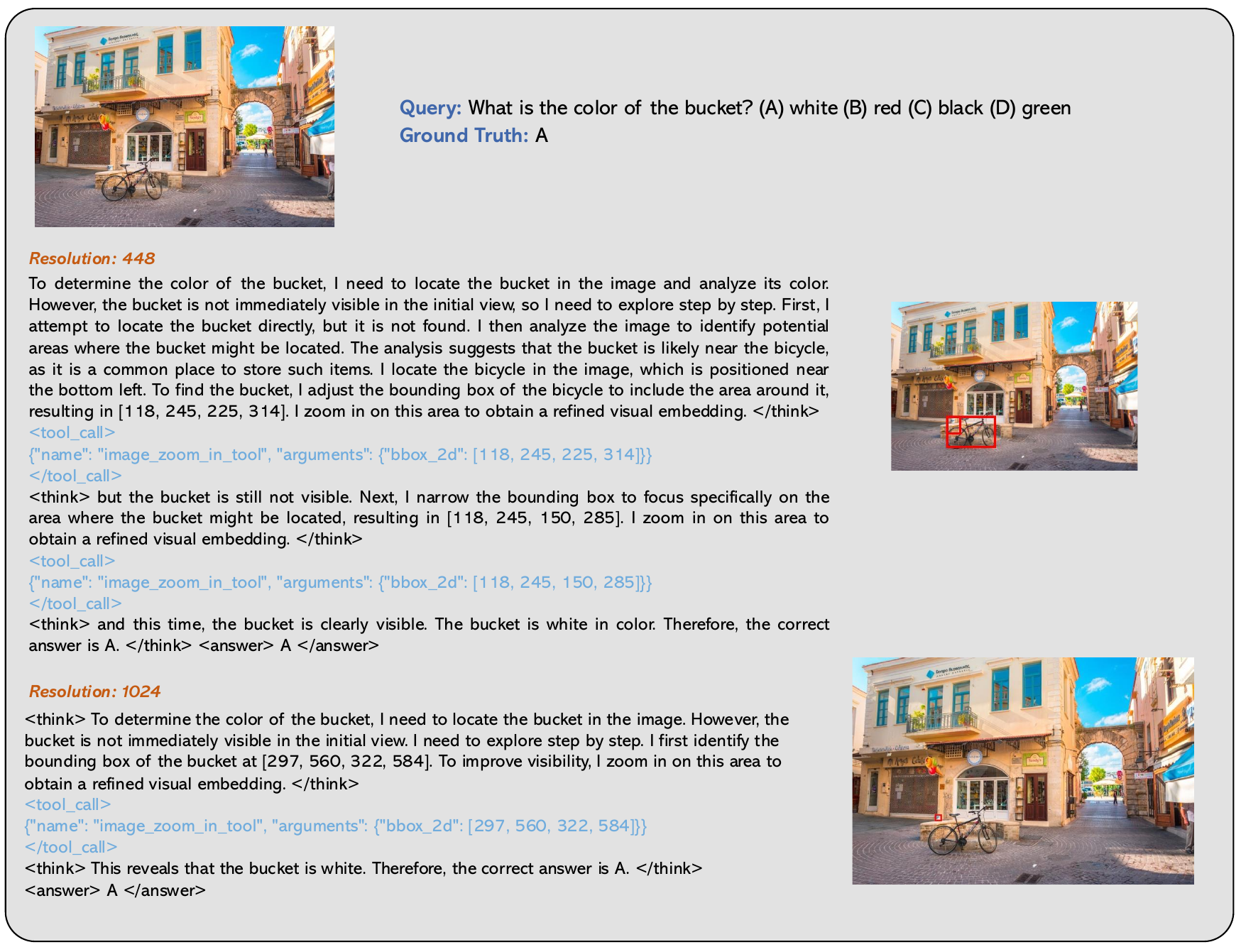}
  \vskip -0.1in
      \caption{Adaptive multimodal reasoning case, Adaptive-CoF adjusts its zoom-in scope based on image clarity at different resolutions.}
      \vskip -0.1in
\label{fig:adaptive_visulization1}
\end{figure}


\begin{figure}[ht]
    \centering
    \includegraphics[width=0.75\textwidth]{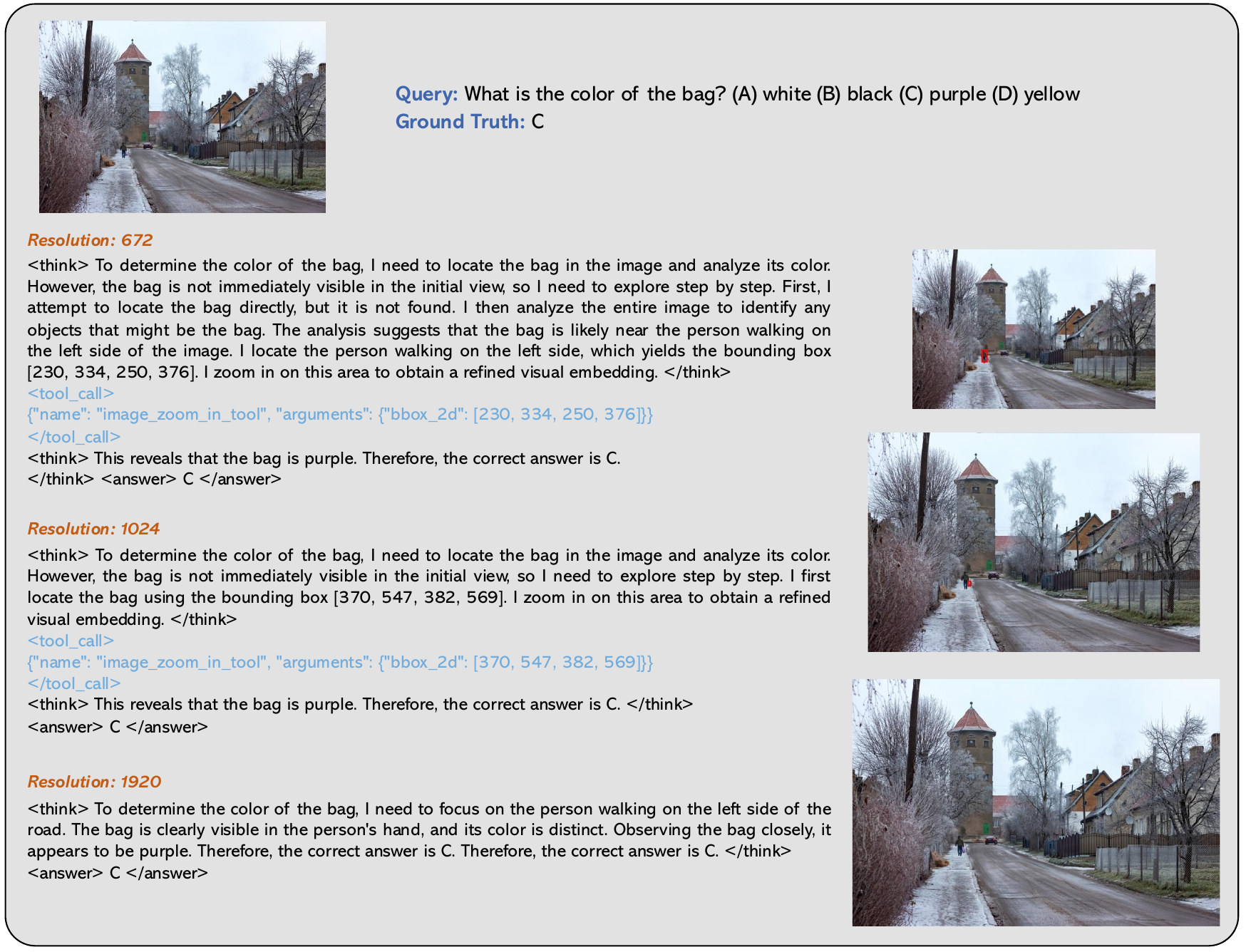}
  \vskip -0.1in
      \caption{Adaptive multimodal reasoning case, Adaptive-CoF transitions from iterative zooming to direct observation as resolution increases.}
      \vskip -0.1in
\label{fig:adaptive_visulization3}
\end{figure}

\subsection{Case Study}

\subsubsection{Visualized Reasoning Cases}

As shown in Figure~\ref{fig:visualization}, Adaptive-CoF adapts its reasoning to the complexity of each visual query. For a simple task like identifying a visible bus, it answers directly, demonstrating efficiency. For a more challenging query, such as recognizing the color of a distant jacket, it performs a single targeted zoom to perceive the answer correctly. Its most advanced behavior appears in an iterative search for a small lock, where it corrects an initial failure before zooming in to locate the object. These examples confirm that Adaptive-CoF enables adaptive reasoning—from simple perception to complex visual search—effectively capturing critical details that would otherwise be missed.

\subsubsection{Adaptive Case Study}

We qualitatively analyze Adaptive-CoF’s adaptive reasoning behavior using two representative examples in Figures~\ref{fig:adaptive_visulization1} and~\ref{fig:adaptive_visulization3}. These examples clearly demonstrate how the model dynamically adjusts its visual search strategy as image resolution varies.

In Figure~\ref{fig:adaptive_visulization1}, the model is asked to identify the color of a bucket. At a low resolution of 448 pixels, the bucket is not visible in the initial image. The model first infers that the bucket may be near the bicycle, a plausible region where such objects appear, and then refines its search through two successive zoom-in operations before locating the bucket and identifying its color as white. At 1024 pixels, the bucket becomes visible but its color remains unclear due to insufficient detail, prompting a single zoom-in for accurate recognition.

A similar behavior is observed in Figure~\ref{fig:adaptive_visulization3}, where the model determines the color of a bag. At 672 pixels, the bag cannot be clearly seen at first. The model hypothesizes that it might be near the person on the left and performs iterative zooming to locate and recognize it. At 1024 pixels, the bag is visible but its color is still ambiguous, so the model conducts one additional zoom-in to confirm the answer. At 1920 pixels, the bag is clearly visible, allowing the model to directly identify its color without zooming.

These examples demonstrate that Adaptive-CoF effectively adjusts its reasoning depth according to visual clarity, transitioning from multi-step exploration at low resolutions to direct answering at high resolutions, thereby maintaining accuracy while reducing computational cost.

\subsection{Ablation Study}

\begin{table}[ht]
\centering
\caption{Ablation Study on MME-RealWorld-Lite and $V^{\star}$.}\vspace{1pt}
\label{tab:ablation}
\resizebox{0.7\textwidth}{!}{
\begin{tabular}{l c c c c c c}
\toprule
\textbf{Model} &
\multicolumn{3}{c}{\textbf{MME-RealWorld}} &
\multicolumn{3}{c}{\textbf{$V^{\star}$ Bench}} \\
\cmidrule[0.2pt](lr){2-4} \cmidrule[0.2pt](lr){5-7}
& Perception & Reasoning & Overall & Attribute & Position & Overall \\
\midrule
Qwen2.5-VL~\cite{bai2025qwen2}  & 46.5& 35.9 & 42.3 & 73.9 & 67.1 & 71.2  \\
\midrule
RL w. Text-only CoT  & 48.6 & 40.3 & 45.4 & 82.6 & 84.2 & 83.3  \\
Cold-start  & 55.2 & 42.3 & 50.1 & 92.2 & 85.5 & 89.5 \\
RL (correctness + format)  & 52.2 & 39.5 & 47.2 & 83.5 & 86.8 & 84.8 \\
\rowcolor{rowgray}
\textbf{Adaptive-CoF}  & \textbf{55.4} & \textbf{44.0} & \textbf{50.9} & \textbf{92.2} & \textbf{86.4} & \textbf{90.1} \\
\bottomrule
\end{tabular}
}
\end{table}

To assess the contribution of each component in our two-stage pipeline and the role of adaptive visual interactions, we conduct an ablation study summarized in Table~\ref{tab:ablation}.

Starting from the Qwen2.5-VL-7B baseline, the supervised fine-tuning (SFT) stage alone (Cold-start) yields a substantial improvement, raising the MME-RealWorld-Lite score from 42.3 to 50.1 and the $V^\star$ Bench score from 71.2 to 89.5. This highlights the SFT stage’s importance in establishing foundational Adaptive-CoF reasoning patterns and tool-calling comprehension.

We further examine an RL variant trained with a reward combining 0.9 correctness and 0.1 format scores (RL (correctness + format)). Although this design aims to balance output correctness and consistency, it proves insufficient to encourage visual reasoning behaviors. As training progresses, the model gradually tends to avoid performing zoom-in reasoning and directly output textual reasoning and final answers, resulting in degraded performance and weaker visual grounding. This observation demonstrates that optimizing only for correctness and format leads to a collapse of adaptive interaction behaviors, emphasizing the necessity of incorporating group-aware reward signals.
Adding the reinforcement learning stage with the adaptive group-aware reward (Adaptive-CoF) provides further refinement, enabling the model to effectively decide when and how to perform zoom-in operations. This adaptive reward formulation better aligns model optimization with the intended visual reasoning objectives.

To test the necessity of explicit visual interaction, we also evaluated an RL w. Text-only CoT variant without zoom-in reasoning. This model showed a clear performance drop, confirming that purely textual reasoning cannot fully substitute adaptive visual search, thus validating our core design.

\section{Conclusion}

In this paper, we present adaptive chain-of-focus (Adaptive-CoF), a framework that enables VLMs to adaptively perform fine-grained visual search and zooming. Through a two-stage training pipeline combining supervised fine-tuning and reinforcement learning, Adaptive-CoF learns to balance detailed perception with computational efficiency, overcoming the trade-off between static viewing and exhaustive zooming. Experiments demonstrate state-of-the-art performance on challenging benchmarks with significantly reduced computational cost.
Although effective, our current work focuses on single-image query–response tasks. Future research will extend this adaptive mechanism to interactive settings such as visual dialogue and multi-image reasoning across scenes ~\cite{zhang2025cross}.
{
\small
\bibliographystyle{plain}
\bibliography{main}
}






\end{document}